\documentclass[letterpaper]{article} 
\usepackage{aaai2026}  
\usepackage{times}  
\usepackage{helvet}  
\usepackage{courier}  
\usepackage[hyphens]{url}  
\usepackage{graphicx} 
\usepackage{natbib}  
\usepackage{caption} 
\usepackage{algorithm}
\usepackage{algorithmic}
\usepackage{xcolor}

\usepackage{newfloat}
\usepackage{listings}
\DeclareCaptionStyle{ruled}{labelfont=normalfont,labelsep=colon,strut=off} 
\floatstyle{ruled}
\newfloat{listing}{tb}{lst}{}
\floatname{listing}{Listing}
\usepackage{graphicx}
\usepackage{amsmath}
\usepackage{amssymb} 
\usepackage{subfigure}
\usepackage{booktabs} 
\usepackage{multirow}
\usepackage{placeins}
\usepackage{afterpage}
\usepackage{caption} 

\usepackage{amsthm}

\theoremstyle{definition}
\newtheorem{definition}{Definition}

\title{Co-EPG: A Framework for Co-Evolution of Planning and Grounding in Autonomous GUI Agents}
\author{
    Yuan Zhao\textsuperscript{\rm 1}{\thanks{The first two authors contributed equally.}},
    Hualei Zhu\textsuperscript{\rm 1}\textsuperscript{*},
    Tingyu Jiang\textsuperscript{\rm 1}{\thanks{Corresponding Author.}},
    Shen Li\textsuperscript{\rm 1}, 
    Xiaohang Xu\textsuperscript{\rm 2}\textsuperscript{†},
    Hao Henry Wang\textsuperscript{\rm 1}
}

\affiliations{
    \textsuperscript{\rm 1}Alibaba Cloud Computing\\
    \textsuperscript{\rm 2}Graduate School of Information Science and Technology, The University of Tokyo\\
    \{zhaoyuan.yz, zhuhualei.zhl, jiangtingyu.jty, lishen.ls,  qiao.wh\}@alibaba-inc.com\\
    xhxu@eidos.ic.i.u-tokyo.ac.jp
}

\usepackage{bibentry}

\begin{document}

\maketitle

\begin{abstract}
Graphical User Interface (GUI) task automation constitutes a critical frontier in artificial intelligence research. While effective GUI agents synergistically integrate planning and grounding capabilities, current methodologies exhibit two fundamental limitations: (1) insufficient exploitation of cross-model synergies, and (2) over-reliance on synthetic data generation without sufficient utilization. To address these challenges, we propose \textbf{Co-EPG}, a self-iterative training framework for \underline{\textbf{Co}}-\underline{\textbf{E}}volution of \underline{\textbf{P}}lanning and \underline{\textbf{G}}rounding. Co-EPG establishes an iterative positive feedback loop: through this loop, the planning model explores superior strategies under grounding-based reward guidance via Group Relative Policy Optimization (GRPO), generating diverse data to optimize the grounding model. Concurrently, the optimized Grounding model provides more effective rewards for subsequent GRPO training of the planning model, fostering continuous improvement. Co-EPG thus enables iterative enhancement of agent capabilities through self-play optimization and training data distillation. On the Multimodal-Mind2Web and AndroidControl benchmarks, our framework outperforms existing state-of-the-art methods after just three iterations without requiring external data. The agent consistently improves with each iteration, demonstrating robust self-enhancement capabilities. This work establishes a novel training paradigm for GUI agents, shifting from isolated optimization to an integrated, self-driven co-evolution approach.

\end{abstract}

\section{Introduction}

In recent years, with the rapid development of large-scale vision language models (LVLMs), building autonomous agents capable of understanding and interacting with graphical user interfaces (GUIs) has emerged as a highly promising application area, attracting extensive research attention ~\cite{OpenAI,li2025websailor}.

The academic community has not yet established a unified paradigm for designing GUI agents. Fundamentally, a GUI agent requires two core capabilities: planning and grounding. Planning determines the action and its values based on the current screen state, while grounding identifies the target element's location. Common approaches involve training a monolithic model~\cite{he2024webvoyager,cheng2024seeclick}, such as directly training on large-scale trajectory data~\cite{xu2024agenttrek,pahuja2025explorer}, pretraining on grounding tasks followed by fine-tuning on planning tasks~\cite{xu2024aguvis,wu2024atlas}, or introducing online Reinforcement Learning (RL) and environment interaction to explore more generalized strategies~\cite{wei2025webagent}. Some studies even employ mechanisms like world model construction~\cite{fang2025webevolver} or self-evolving curriculum learning~\cite{qi2024webrl} to enable continuous self-improvement of model capabilities. However, these monolithic models increasingly expose limitations in perception and interaction when generalized to diverse GUI environments~\cite{gou2024navigating}. Consequently, the research focus has shifted towards more flexible modular designs, primarily characterized by task decoupling and multi-model collaboration. For instance, some studies employ a collaborative framework between high-level planning and low-level grounding to enhance the accuracy and flexibility of agents ~\cite{zhang2025litewebagent}, while others construct cooperative multi-agent systems to better handle complex tasks~\cite{zhao2025cola,zhang2024ufo,wang2024mobile,liu2025infiguiagent}. However, current collaborative architectures for GUI agents face two critical challenges: (1) They predominantly rely on independent model optimization, which neglects the potential for synergistic co-evolution between interdependent components like planning and grounding. (2) This reliance fosters a dependency on vast synthetic datasets, underutilizing existing data and introducing synthetic noise. Therefore, it is imperative to develop a novel collaborative paradigm that enables the synergistic evolution of planning and grounding models while maximizing the utility of available data.

In this study, we propose Co-EPG, a self-iterative training framework for Co-Evolution of Planning and Grounding. The core of Co-EPG is a positive feedback loop that drives the co-evolution of both models. The planning model explores new strategies using GRPO to generate diverse and specific plans, which progressively enhance the grounding model's execution capabilities. In turn, the improved grounding model delivers higher-quality rewards, guiding the planning model toward more effective strategies. This iterative loop, where the plan bridges the two models and provides specific information to the grounding model, enables continuous self-improvement. During the collaborative training of both models, we propose a confidence-based dynamic reward ensemble mechanism (C-DREM), which effectively reduces reward noise by aggregating rewards from multiple grounding models with confidence-based weighting. Our contributions can be summarized as follows:

\begin{enumerate}
    \item We propose Co-EPG, a self-iterative training framework for Co-Evolution of Planning and Grounding. The framework establishes a positive feedback loop in which the grounding model guides the planning model's strategy exploration through reward, while the optimized planning model generates high-quality data to further enhance the grounding model. This closed-loop self-improvement mechanism drives the continuous co-evolution of both models.
    
    \item We present the C-DREM that harnesses multiple grounding models to assess plan executability. By dynamically weighting reward based on each model’s confidence score, C-DREM constructs a robust composite reward signal, which significantly enhances the stability and accelerates the convergence of the GRPO training process.

    \item The experimental results indicate that the Co-EPG exhibits excellent generalization by solely relying on the benchmark dataset for self-iterative optimization. It outperforms existing state-of-the-art methods on both Multimodal-Mind2web (58.4\%) and AndroidControl (83.1\%) benchmarks.
\end{enumerate}

\section{Related Work}
Recent advances in Large Language Models (LLMs) and Vision Language Models (VLMs) have laid a solid foundation for research on GUI agents~\cite{liu2024deepseek,achiam2023gpt,wang2024qwen2,bai2023qwen}. Current researches primarily revolve around two core views: architecture design and capability enhancement.

\subsection{GUI Agent Architecture Design}
GUI agent architectures have evolved from LLM-centric frameworks to end-to-end VLM-driven agents, and most recently, flexible modular systems. Early agents are built around LLMs ~\cite{gur2023real,zhao2024expel,fu2024autoguide,wang2024large}, but they struggle with complex visual information. This requires auxiliary models for tasks like element selection~\cite{deng2023mind2web} or information parsing~\cite{lee2025learning}. However, this reliance on text overlooks crucial visual and semantic cues~\cite{zheng2024gpt}. Consequently, the field pivots towards VLMs to natively integrate visual data. These fall into two main categories: hybrid approaches~\cite{yang2023set,lu2024omniparser}, which use external tools to parse screen information for VLMs; and end-to-end VLM-driven agents~\cite{niu2024screenagent,qin2025ui,he2024webvoyager,cheng2024seeclick}.
However, these models face new generalization challenges in diverse GUI environments~\cite{gou2024navigating}. To overcome generalization issues, recent works focus on flexible, modular designs, which involve either decoupling planning and grounding for accuracy and flexibility~\cite{zheng2024gpt,zhang2025litewebagent} or employing multi-agent collaboration for complex tasks~\cite{liu2025infiguiagent,agashe2025agent}.

\subsection{GUI Agent Capability Enhancement}
To enhance GUI agent capabilities, researches focus on two main fronts: data synthesis and training strategies. 

\noindent\textbf{Automated Data Synthesis.}
Data synthesis often depends on significant human labor, leading to high costs. To replace costly manual annotation, various automatic methods have been proposed. Explorer ~\cite{pahuja2025explorer} generates over 94K successful trajectories via dynamic exploration of web environments. AgentTrek~\cite{xu2024agenttrek} simulates execution traces using web tutorials as step-by-step guides. Winclick~\cite{hui2025winclick} builds a 60k-sample dataset by identifying interactive elements from raw screenshots and generating corresponding natural language instructions. To improve data quality, Aguvis~\cite{xu2024aguvis} generates ``inner monologues'' to enhance logical reasoning.

\noindent\textbf{Advanced Training Strategies.}
Researchers have explored two primary paths for training. 
The first involves optimizing Supervised Fine-Tuning (SFT), such as two-stage training~\cite{xu2024aguvis,wu2024atlas} and curriculum learning~\cite{chen2025guicourse}. However, the generalization capability of SFT-based agents is heavily dependent on the scale of the training data \cite{ jiang2025importanceawaredataselectionefficient}. Additionally, AgentSymbiotic~\cite{zhang2025symbiotic} proposes collaborative learning between large and small LLMs, though its advantages have not yet been applied to decoupled planning-and-grounding architectures. The second path leverages RL, such as rule-based rewards~\cite{luo2025gui,wei2025webagent} or hybrid distillation-RL approaches~\cite{liu2025infigui}. More advanced agents aim to self-evolve by generating tasks from failures~\cite{qi2024webrl}, using co-evolving world models~\cite{fang2025webevolver}, and applying attention-guided self-improvement ~\cite{yuan2025enhancing}. These ideas are largely confined to end-to-end models and have not been integrated with modular architectures. \textit{To the best of our knowledge, our work is the first to propose a co-evolution framework for decoupled architectures that iteratively refines planning and grounding capabilities to maximize data value.}

\begin{figure*}[!ht]
    \centering
    \includegraphics[width=\textwidth]{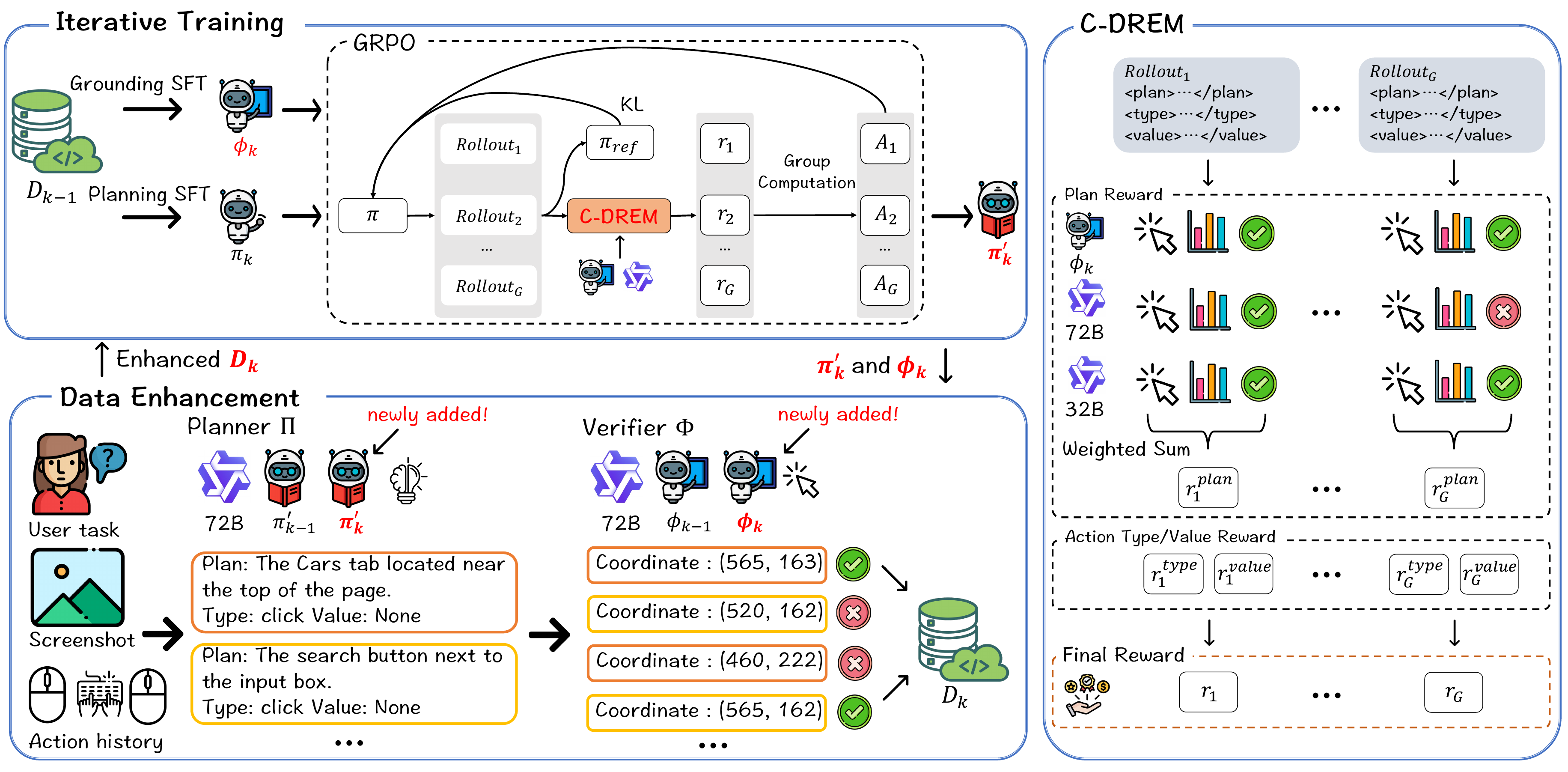}
    \caption{Overview of our proposed Co-EPG framework. The framework drives the co-evolution of the planning model ($\pi$) and the grounding model ($\phi$) through an optimization loop. This loop alternates between Iterative Training, which employs C-DREM to drive the co-evolution of both models, and Data Enhancement, which refines the dataset for the next iteration.}
    \label{fig:overview}
\end{figure*}
\section{Preliminaries}
We formulate the GUI task as a Partially Observable Markov Decision Process (POMDP) following ~\cite{xu2024aguvis}. 

\begin{definition}[POMDP]
Formally, a POMDP is defined by a tuple ($\mathcal{S}$, $\mathcal{A}$, $\mathcal{O}$, $\mathcal{F}$, $\mathcal{R}$), where $\mathcal{S}$ represents the complete state space, $\mathcal{O}$ is the set of partial observations the agent can perceive (e.g., visual screenshots and HTML content), and $\mathcal{A}$ denotes the space of feasible actions. $\mathcal{F}$ represents the state transition function, $\mathcal{F}(s, a, s')$ = $P(s'|s,a)$, where is the probability of transitioning to state $s'$ after taking action $a \in \mathcal{A}$ in state $s \in \mathcal{S}$. $\mathcal{R}$ is the reward function, $\mathcal{S} \times \mathcal{A} \rightarrow [0,1]$, which specifies the reward for taking action $a$ in state $s$. At each timestep $t$, the agent determines an action based on the current observation $o_t=(o_t^{vision},o_t^{html}) \in \mathcal{O}$, the task description $Q$, and its interaction history $h_t = \{a_0, a_1, ..., a_{t-1}\}$. The action $a_t$ is a composite tuple ($a^{coor}_t$, $a^{type}_t$, $a^{value}_t$), where $a^{coor}_t$ represents the target coordinates of action, $a^{type}_t$ is the action type, and $a^{value}_t$ is the action value. This process generates a complete interaction trajectory $T$ = \{($o_1$, $a_1$), \ldots, ($o_{|T|}$, $a_{|T|}$)\}, where $|T|$ is the total number of steps.
\end{definition}

\section{Methodology}
In this section, we propose a self-iterative training framework for \underline{\textbf{Co}}-\underline{\textbf{E}}volution of \underline{\textbf{P}}lanning and \underline{\textbf{G}}rounding (\textbf{Co-EPG}), built upon the planning model $\pi$ and the grounding model $\phi$, as illustrated in Figure~\ref{fig:overview}.

\subsection{P-G Dual-Model}
Inspired by ~\cite{gou2024navigating}, we adopt a P-G dual-model architecture whose decoupled design allows each model to specialize in its respective function, thereby enabling the GUI agent to efficiently manage complex multi-step tasks. At each timestep $t$, the complex decision-making process is divided into two cooperative subtasks. Specifically, the planning model $\pi$ acts as a high-level strategist. Given the current observation $o_t$, task description $Q$, and interaction history $h_t$, it generates a multi-part action decision, which comprises a textual plan $p_t$, an action type $a^{type}_t$, and a corresponding action value $a^{value}_t$, as follows:

\begin{equation}
p_t, a^{type}_t, a^{value}_t = \pi(Q, o_t, h_t).   
\end{equation}

Subsequently, the grounding model $\phi$ utilizes the specific plan $p_t$ with the visual input from the vision observation (e.g., a screenshot) $o_t^{vision}$ to predict the exact coordinates $a^{coor}_t$ of the target element, as follows:
\begin{equation}
a^{coor}_t = \phi(o_t^{vision}, p_t).
\end{equation}
The outputs of these two models are then combined to form the current action $a_t = (a^{coor}_t, a^{type}_t, a^{value}_t)$ to be executed. This sequence of operations is repeated iteratively, generating a complete task-execution trajectory $T$.

\subsection{Co-Evolving Optimization Loop}
\label{Co-Evolving Optimization Loop}
The self-iterative collaborative training loop of Co-EPG primarily includes the following two core steps: Iterative Training and Data Enhancement.

\noindent\textbf{Iterative Training.} Training is driven by the iterative dataset $D_{k}$. In each iteration $k$ (where $k\geq1$), we first fine-tune the model on the dataset $D_{k-1}$ to obtain $\pi_{k}$ and $\phi_{k}$. Then, we refine the planning model $\pi_{k}$ into $\pi'_{k}$ through collaborative GRPO training, which is guided by our proposed C-DREM. The mechanism adaptively aggregates the rewards generated from an ensemble of the grounding model $\phi_{k}$ and VLMs (e.g., Qwen2.5-VL-72B-Instruct and Qwen2.5-VL-32B-Instruct), which is detailed further in Section~\ref{sec:C-DREM}. Guided by these grounding models, the planning model explores more successful strategies, which makes its plans more comprehensible. The grounding model $\phi_{k}$, in turn, is further strengthened by fine-tuning on high-quality data distilled from the previous stage, which enhances its perception capabilities. This process iteratively aligns the capabilities of the planning and grounding models, ultimately yielding the planning model $\pi'_{k}$ and the grounding model $\phi_{k}$.

\noindent\textbf{Data Enhancement.} We develop a self-enhancing data evolution mechanism: Initially ($k=0$), we form two specialized pools, each consisting of open-source VLMs: the Planner $\Pi$ and the Verifier $\Phi$. The Planner $\Pi$ generates the specific plan $p_t$ based on the current observation $o_t$, task description $Q$, and historical actions $h_t$. The Verifier 
$\Phi$ then validates each plan and constructs the initial dataset $(o_t, h_t, p_t, a_t) \in D_0$ by retaining only successfully verified plans. In subsequent iterations ($k\geq1$), the updated planning model $\pi'_{k}$ and grounding model $\phi_{k}$ participate in the data production process: the Planner $\Pi$ incorporates $\pi'_{k}$ to enhance planning diversity, and the Verifier $\Phi$ integrates $\phi_{k}$ to improve discrimination reliability. To balance effectiveness and efficiency, only the latest $\{\pi'_{k},\pi'_{k-1}\}$ and $\{\phi_{k},\phi_{k-1}\}$ are reserved for data production, maintaining the size of both pools. Through a self-evolution loop, the planning and grounding models achieve synergistic improvement. Ultimately, we obtain a more powerful GUI agent $M_k$= $\{\pi'_{k}, \phi_{k}\}$.

\subsection{C-DREM} \label{sec:C-DREM}
As described in Section~\ref{Co-Evolving Optimization Loop}, we facilitate synergy between the planning and grounding models through collaborative GRPO training. To build a more comprehensive reward signal ~\cite{liu2025structural}, we evaluate the planning model's output on its plan, action type, and action value. For clarity, we omit the time step subscript $t$ in the following contents, as all formulas are defined under a unified time step.

\noindent\textbf{Plan Reward.}
A key challenge for the planning model during GRPO training is that the quality of a generated plan cannot be directly evaluated, because its effectiveness is ultimately determined by whether the grounding model can use this plan to accurately locate the target element. As a result, we use the grounding model's prediction accuracy as a reward to guide the optimization of the planning model. The accuracy, denoted as ${Acc}^{plan}$, is calculated based on whether the coordinates $a^{coor}$ predicted by the grounding model fall within the target's bounding box $bbox$:

\begin{equation}
{Acc}^{plan} = 
\begin{cases} 
1, & \begin{array}{@{}l@{}} 
        \text{if } a^{coor} \in bbox , 
    \end{array} \\
0, & \text{otherwise.}
\end{cases}
\end{equation}

To address the inherent bias and poor performance of single reward model on out-of-distribution data, we propose C-DREM, a confidence-based dynamic reward ensemble mechanism. The core idea of C-DREM is to aggregate the collective intelligence of diverse grounding models, which include both open-source models and the grounding model $\phi_k$ generated during our iterative training process. The reward $r^{plan}$ is defined as a weighted sum:
\begin{equation}
r^{plan} = \sum_{j=1}^N w_j \cdot {Acc}^{plan}_{j},
\end{equation}
where ${Acc}^{plan}_{j}$ represents the reward from the $j$-th grounding model (for $j = 1, \ldots, N$), $N$ is the number of grounding models, and $w_j$ is the corresponding weight.
The weight $w_j$ is determined by two components: a static prior $\sigma_j$ and a dynamic confidence score $c_{j}$:
\begin{equation}
w_j = \frac{\exp(\sigma_j \cdot c_j)}
{\sum_{n=1}^N \exp(\sigma_n \cdot c_n)}.
\end{equation}
Here, the static prior $\sigma_j$ is set higher for our trained grounding model, reflecting its critical importance in production deployment. The dynamic confidence score $c_{j}$ is calculated as the sum of the log-likelihoods of the predicted coordinate token $\tau$ of $a_j^{coor}$, normalized by its length $L$: 

\begin{equation}
c_j = \frac{1}{L} \sum^{L}_{l=1} \log P(\tau_l \mid o, h, p).
\end{equation}

Next, we calculate rewards based on action type and action value according to ~\cite{luo2025gui}. 

\noindent\textbf{Action Type Reward.} The action type reward ${r}^{type}$ depends on whether the predicted action type $a^{type}$ exactly matches the ground truth action type $gt^{type}$:
\begin{equation}
{r}^{type} =
\begin{cases}
1, & \text{if } a^{type} = gt^{type}, \\
0, & \text{otherwise}.
\end{cases} 
\end{equation}
\noindent\textbf{Action Value Reward.} The action value reward ${r}^{value}$ is calculated based on the $F_1$ score between the predicted value $a^{value}$ and the ground truth value $gt^{value}$, which is called as:

\begin{equation}
{r}^{value} = 
\begin{cases}
1, & \text{if } F_1(a^{value},gt^{value}) > 0.5, \\
0, & \text{otherwise}.
\end{cases}
\end{equation}

\noindent\textbf{Final Reward.} The final reward $r_i$ of the $i$-th generation for the GUI tasks is calculated by ${r}^{plan}_i$, ${r}^{type}_i$ and ${{r}^{value}_i}$:

\begin{equation}
r_i = 
\begin{cases}
0, & \text{if } r^{type}_i=0 \text{ or } r^{value}_i=0, \\
r_i^{plan}, & \text{otherwise}.
\end{cases}
\end{equation}

\noindent\textbf{Group Computation.} Subsequently, we normalize these rewards across the $G$ generated $Rollouts$ to compute the advantage $A_i$, which subsequently serves as the objective for GRPO policy optimization~\cite{shao2024deepseekmath}:
\begin{equation}
A_i = \frac{r_i - mean(\{r_1, r_2, \cdots, r_G\})}{std(\{r_1, r_2, \cdots, r_G\})}.
\end{equation}

\begin{table*}[!ht]
    \centering
    \resizebox{\textwidth}{!}
    {
    \begin{tabular}{llcccccccccc}
    \toprule
    \multirow{2}{*}{\textbf{Planner}} & \multirow{2}{*}{\textbf{Grounder}} & \multicolumn{3}{c}{\textbf{Cross-Task}} & \multicolumn{3}{c}{\textbf{Cross-Website}} & \multicolumn{3}{c}{\textbf{Cross-Domain}} & \multirow{2}{*}{\textbf{Avg SR}} \\
    \cmidrule(lr){3-5} \cmidrule(lr){6-8} \cmidrule(lr){9-11}
    & & \textbf{Ele.Acc} & \textbf{Op.F1} & \textbf{Step SR} & \textbf{Ele.Acc} & \textbf{Op.F1} & \textbf{Step SR} & \textbf{Ele.Acc} & \textbf{Op.F1} & \textbf{Step SR} \\
    \midrule
    \multirow{3}{*}{GPT-4} & Choice & 46.4 & 73.4 & 40.2 & 38.0 & 67.8 & 32.4 & 42.4 & 69.3 & 36.8 & 36.5 \\
     & SoM & 29.6 & - & 20.3 & 20.1 & - & 13.9 & 27.0 & - & 23.7 & 19.3 \\ 
    \cmidrule(lr){1-12}
    \multirow{3}{*}{GPT-4o} & SeeClick & 32.1 & - & - & 33.1 & - & - & 33.5 & - & - & - \\
    & UGround-V1-2B & 48.6 & - & - & 47.6 & - & - & 47.7 & - & - & - \\ 
    & UGround-V1-7B & 50.7 & - & - & 48.1 & - & - & 48.5 & - & - & - \\ 
    \cmidrule(lr){1-12}
    GPT-4V & OmniParser & 42.4 & 87.6 & 39.4 & 41.0 & 84.8 & 36.5 & 45.5 & 85.7 & 42.0 & 39.3 \\
    \cmidrule(lr){1-12}
    \multicolumn{2}{c}{Explorer-4B} & 53.4 & 88.1 & 50.7 & 55.6 & 89.5 & 51.4 & 49.8 & 88.8 & 47.2 & 49.8 \\
    \multicolumn{2}{c}{Explorer-7B} & 56.5 & \underline{90.3} & 53.2 & 60.5 & \underline{90.7} & \underline{56.7} & 55.7 & \underline{90.4} & 53.0 & 54.3 \\
    \multicolumn{2}{c}{AgentTrek-7B} & 60.8 & 88.9 & 55.7 & 57.6 & 88.1 & 51.4 & 56.0 & 87.5 & 52.6 & 53.2 \\
    \multicolumn{2}{c}{SeeClick-9.6B} & 28.3 & 87.0 & 25.5 & 21.4 & 80.6 & 16.4 & 23.2 & 84.8 &  20.8 & 20.9 \\
    \multicolumn{2}{c}{AGUVIS-7B} & \underline{64.2} & 89.8 & \underline{60.4} & \underline{60.7} & 88.1 & 54.6 & \textbf{60.4} & 89.2 & \textbf{56.6} & \underline{57.2} \\
    \midrule
    \multicolumn{2}{c}{Co-EPG-Web-3B} &  57.7 & 88.8 & 53.1 & 56.9 & 87.7 & 51.1 & 53.6 & 89.6 & 50.0 & 51.4  \\
    \multicolumn{2}{c}{Co-EPG-Web-7B} & \textbf{66.3} & \textbf{92.4} & \textbf{61.9} & \textbf{62.3} & \textbf{91.7} & \textbf{58.1} & \underline{59.3} & \textbf{92.2} & \underline{55.3} & \textbf{58.4} \\
    \bottomrule
    \end{tabular}
    }
    \caption{Performance on Multimodal-Mind2Web. We bold the best results and underline the second-best performance.}
    \label{tab:mm_mind2web_main}
\end{table*}

\section{Experiments}
\label{sec:Experiments}

\subsection{Datasets and Metrics}
We conduct systematic evaluations on two main benchmarks: Multimodal-Mind2Web~\cite{deng2023mind2web} for web interactions and AndroidControl~\cite{li2024effects} for mobile applications. For Multimodal-Mind2Web, we report element accuracy (Ele.Acc), operation F1 (Op.F1), and step success rate (Step SR), while for AndroidControl, we use the standard metric of step accuracy (Step Acc). Detailed benchmark information is provided in Appendix. A.

\label{sec:dataset}
\subsection{Implementation Details} 
We implement the experiments using PyTorch 2.6.0 on a Linux server equipped with 984GB RAM, Intel Xeon Platinum 8369B CPU @ 2.90GHz, and Nvidia A100 Tensor Core 80GB GPUs. We select Qwen2.5-VL as our backbone in experiments, with the training pipeline implemented based on the MS-SWIFT framework. All results represent the average of three independent runs to ensure credibility. Training details are provided in Appendix. B.

\subsection{Main Results}

\noindent\textbf{Multimodal-Mind2Web Results.} The experimental results in Table~\ref{tab:mm_mind2web_main} highlight the self-improvement capability of our proposed Co-EPG framework, which achieves state-of-the-art performance on web tasks. Specifically, the Co-EPG-Web-7B model achieves 58.4\% on average Step SR of three subtasks, outperforming both Explorer-7B~\cite{pahuja2025explorer} (54.3\%) and the previous leading model, AGUVIS-7B~\cite{xu2024aguvis} (57.2\%). Notably, unlike AGUVIS-7B, which relies on extensive auxiliary data construction, Co-EPG achieves superior performance using only the original benchmark data. Co-EPG also shows strong generalization at smaller scales. Co-EPG-Web-3B outperforms the previous state-of-the-art model with similar parameters, Explorer-4B~\cite{pahuja2025explorer}, by nearly 1.6\% on average Step SR. Based on the experimental results across varying model scales, this consistent performance gain demonstrates that the co-evolution of planning and grounding models enables agents to learn from more diverse and higher-quality data, without requiring external data.

\begin{table}[!ht]
\centering
\small
    \begin{tabular}{ll ccc c}
    \toprule
    \multirow{2}{*}{\textbf{Planner}} & \multirow{2}{*}{\textbf{Grounder}} & \multicolumn{2}{c}{\textbf{Step Acc}} & \multirow{2}{*}{\textbf{Avg Acc}} \\
    \cmidrule(lr){3-4} 
    & & \textbf{High} & \textbf{Low} & \\
    \midrule
    GPT-4o & SeeClick & 41.8 & 52.8 & 47.3 \\
    GPT-4o & UGround-v1-2B & 50.0 & 65.0 & 57.5 \\
    GPT-4o & UGround-v1-7B & 49.8 & 66.2 & 58.0 \\
    \midrule
    \multicolumn{2}{c}{ AGUVIS-7B } & 61.5 & 80.5 & 71.0 \\
    \multicolumn{2}{c}{ AGUVIS-72B } & 66.4 & 84.4 & 75.4 \\
    \multicolumn{2}{c}{ OS-Atlas-4B } & 67.5 & 80.6 & 74.1 \\
    \multicolumn{2}{c}{ OS-Atlas-7B } & 71.2 & 85.2 & 78.2  \\
    \multicolumn{2}{c}{ GUI-R1-3B } & 46.6 & 64.4 & 55.5 \\
    \multicolumn{2}{c}{ GUI-R1-7B } & 51.7 & 66.5 & 59.1  \\
    \multicolumn{2}{c}{ UI-R1-3B } & 45.4 & 66.4 & 55.9 \\
    \multicolumn{2}{c}{ UI-TARS-2B } & 68.9 & 89.3 & 79.1 \\
    \multicolumn{2}{c}{ UI-TARS-7B } & 72.5 & 90.8 & 81.7 \\
    \multicolumn{2}{c}{ InfiGUI-R1-3B } & 71.1 & \textbf{92.1} & 81.6 \\
    \midrule
    \multicolumn{2}{c}{Co-EPG-Mob-3B} & \underline{73.4} & 90.2 & \underline{81.8}  \\ 
    \multicolumn{2}{c}{Co-EPG-Mob-7B} & \textbf{74.2} & \underline{92.0} & \textbf{83.1}  \\  
    \bottomrule
    \end{tabular}
    \caption{Performance on AndroidControl. We bold the best results and underline the second-best performance.
    }
    \label{androidcontrol_main}
\end{table}

\noindent\textbf{AndroidControl Results.} 
Table~\ref{androidcontrol_main} presents the experimental results of Co-EPG on the AndroidControl benchmark for the mobile task. Co-EPG-Mob-7B achieves the best performance with 83.1\% on average Step Acc of high-level and low-level tasks, demonstrating a significant 1.4\% advantage over the previous state-of-the-art UI-TARS-7B~\cite{qin2501ui}. This strong performance extends to smaller models as well: the Co-EPG-Mob-3B variant also performs excellently with a score of 81.8\%, maintaining a competitive advantage compared to InfiGUI-R1-3B~\cite{liu2025infigui}   (81.6\%). These experiments validate the effectiveness and generalization of Co-EPG across diverse GUI environments.

\subsection{Ablation Study}
The effectiveness of the Co-EPG framework arises from three synergistic design principles: the P-G dual-model architecture, the self-iterative loop for continuous evolution, and the confidence-based dynamic reward ensemble mechanism (C-DREM) for precise and adaptive guidance. In this section, we conduct three iterations and use $M_k$ (detailed in Section~\ref{Co-Evolving Optimization Loop}) to represent the agent after each iteration.

\subsubsection{Impact of P-G Dual-Model.}
\begin{table}[!ht]
    \centering
    \small
    \begin{tabular}{lccc}
        \toprule
        Method & Avg SR \\
        \midrule
        End2End & 50.1 \\
        Co-EPG-Web-7B-$\text{M}_1$  & \textbf{53.5} \\
        \bottomrule
    \end{tabular}
    \caption{Performance Comparison: Decoupled P-G Dual-Model Architecture vs. End-to-End.} 
    \label{tab:table_2} 
\end{table}

We validate the contributions of the P-G dual-model by analyzing model variants with and without the decoupling structure mediated by planning instructions. Specifically, the performance of the P-G dual-model is compared with that of an end-to-end model under fine-tuning training. The experimental results, as shown in Table~\ref{tab:table_2}, demonstrate that the decoupled architecture improves performance by 3.4\% over the end-to-end approach, confirming its effectiveness.

\begin{figure}[!t]
    \centering
\includegraphics[width=0.9\columnwidth]{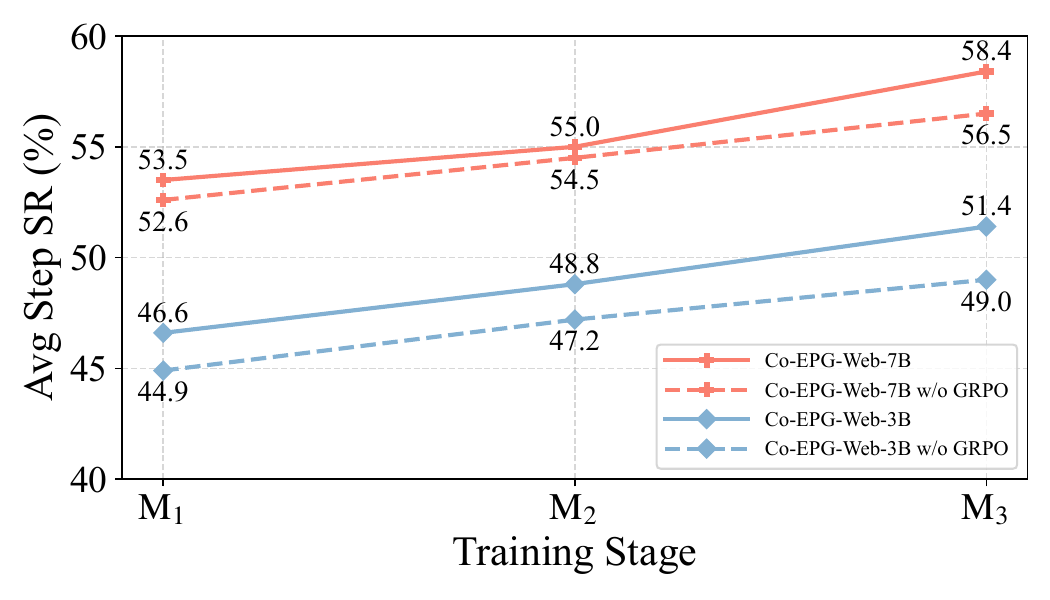}
    \caption{Effectiveness of the self-iterative evolution mechanism in Co-EPG on Multimodal-Mind2Web.}
    \label{fig:experiment_2_1}
\end{figure}

\noindent\textbf{Impact of Iterative Evolution.}
To evaluate the synergistic co-evolution of Co-EPG, we track the performance of Co-EPG-Web-3B/7B across multiple iterations. As illustrated in Figure~\ref{fig:experiment_2_1}, both models show clear and steady improvements, confirming the performance gains from our self-evolutionary framework. To determine whether this gain is from data iteration alone or whether GRPO collaborative training also plays a crucial role, we remove GRPO and rely solely on SFT. Although the models without GRPO (dashed lines) also improve, they significantly underperform the full Co-EPG architecture. This result demonstrates that data iteration and GRPO collaborative training are dual drivers of this evolution. Specifically, GRPO collaborative training acts as an accelerator, providing the precise exploration rewards required to transcend the performance limitations inherent to purely data-centric refinement.

\begin{table}[!ht]
    \centering
    \small
    \begin{tabular}{lccc}
        \toprule
        Method & Avg SR \\
        \midrule       
        w/o C-DREM & 56.50 \\
        w/o Confidence \& Prior Weights  & 57.01 \\
        w/o Confidence Weights & 57.67 \\
        Co-EPG-Web-7B-$\text{M}_3$ & \textbf{58.41} \\
        \bottomrule
    \end{tabular}
    \caption{Comparative Analysis of C-DREM in GRPO Collaborative Training.} 
    \label{tab:table_3} 
\end{table}
\noindent\textbf{Impact of C-DREM.} To validate the effectiveness of key components in C-DREM, we conduct ablation studies with the following variants:

\begin{itemize}
    \item \textbf{w/o C-DREM}. Uses the trained grounding model as the single reward model.
    \item \textbf{w/o Confidence \& Prior Weights}. Uses average weighting but removes confidence and prior weights.
    \item \textbf{w/o Confidence Weights}. Uses prior weighting but removes the confidence weights.
\end{itemize}

We establish a performance baseline by using a single grounding model as the reward model (w/o C-DREM). Table~\ref{tab:table_3} demonstrates three key findings: Firstly, average weighting (w/o Confidence \& Prior Weights) already shows 0.51\% improvement over the baseline. Secondly, prior weighting (w/o Confidence Weights) achieves a further 0.66\% gain but with static weights. These verify the ensemble's inherent advantage in generating robust rewards. Finally, our complete C-DREM achieves a 1.91\% improvement, highlighting the critical role of confidence-based dynamic weighting. These findings confirm that C-DREM can generate more adaptive and precise rewards, addressing the limitations of static weights, and collectively validate the effectiveness of C-DREM in improving model performance.

\subsection{Case Study}
\begin{figure}[!ht]
    \centering
    \includegraphics[width=\columnwidth]{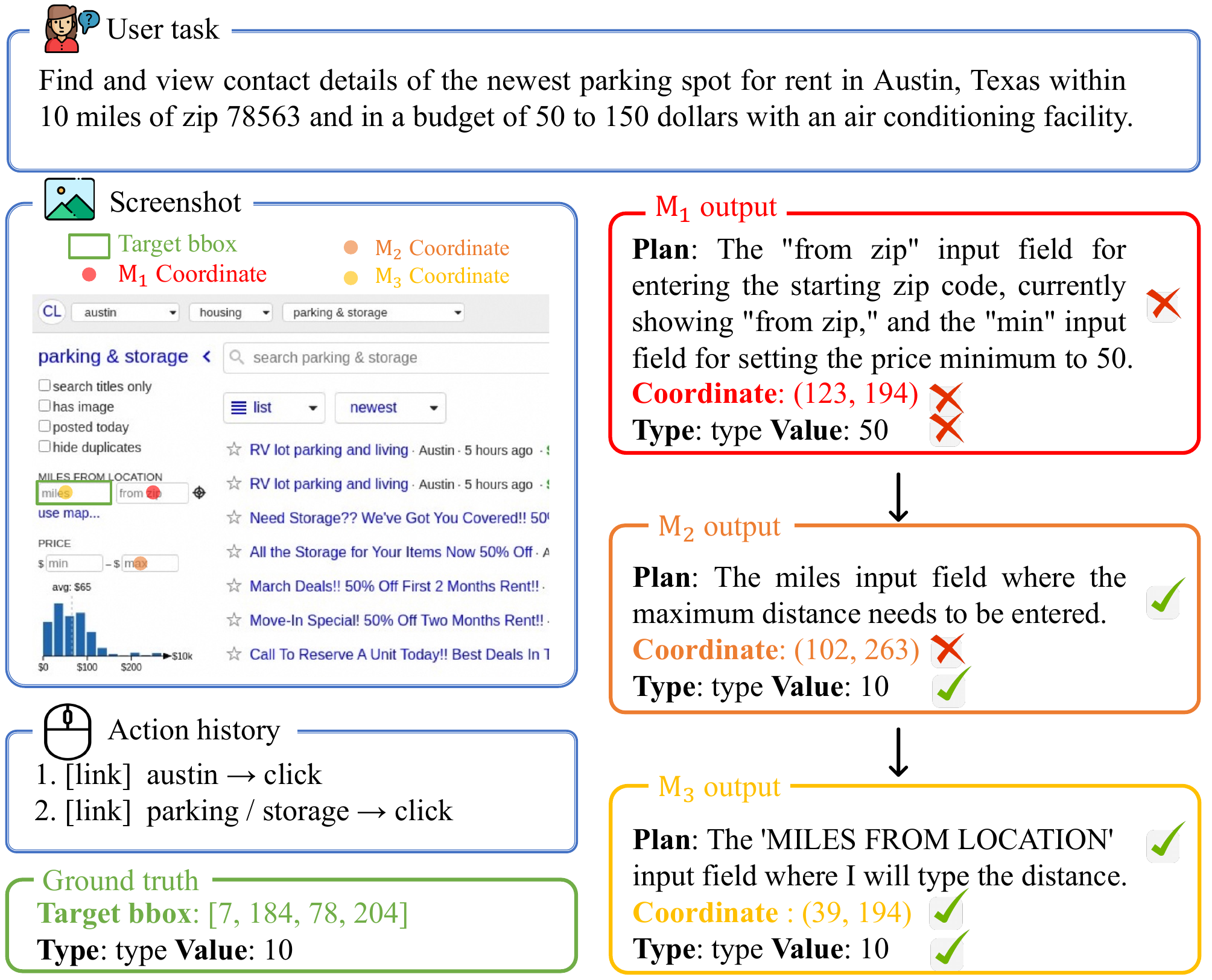}
    \caption{An Evolution Example of Co-EPG.
    }
    \label{fig:experiment_case}
\end{figure}
\noindent As shown in Figure~\ref{fig:experiment_case}, we visualize evolution in the capabilities of the planning and grounding models across three stages. In $M_{1}$ stage, the planning model not only produces an ambiguous plan by merging multiple UI targets but also predicts an incorrect action value. In $M_{2}$ stage, the planning model is able to identify the correct UI element and action value for the task, however, due to the misalignment between the capabilities of the planning and grounding model, the predicted coordinates are still incorrect. Moreover, in $M_{3}$ stage, the planning model achieves high semantic accuracy by generating plans with specific UI text ("MILES FROM LOCATION") rather than generic descriptions ("miles input field"), leading to comprehensive success. This progression from ambiguous to precise planning, along with the improved capabilities of the grounding model, greatly boosts the model’s overall performance on GUI tasks.
\subsection{Efficiency Study}
\begin{figure}[ht]
    \centering
\includegraphics[width=0.9\columnwidth]{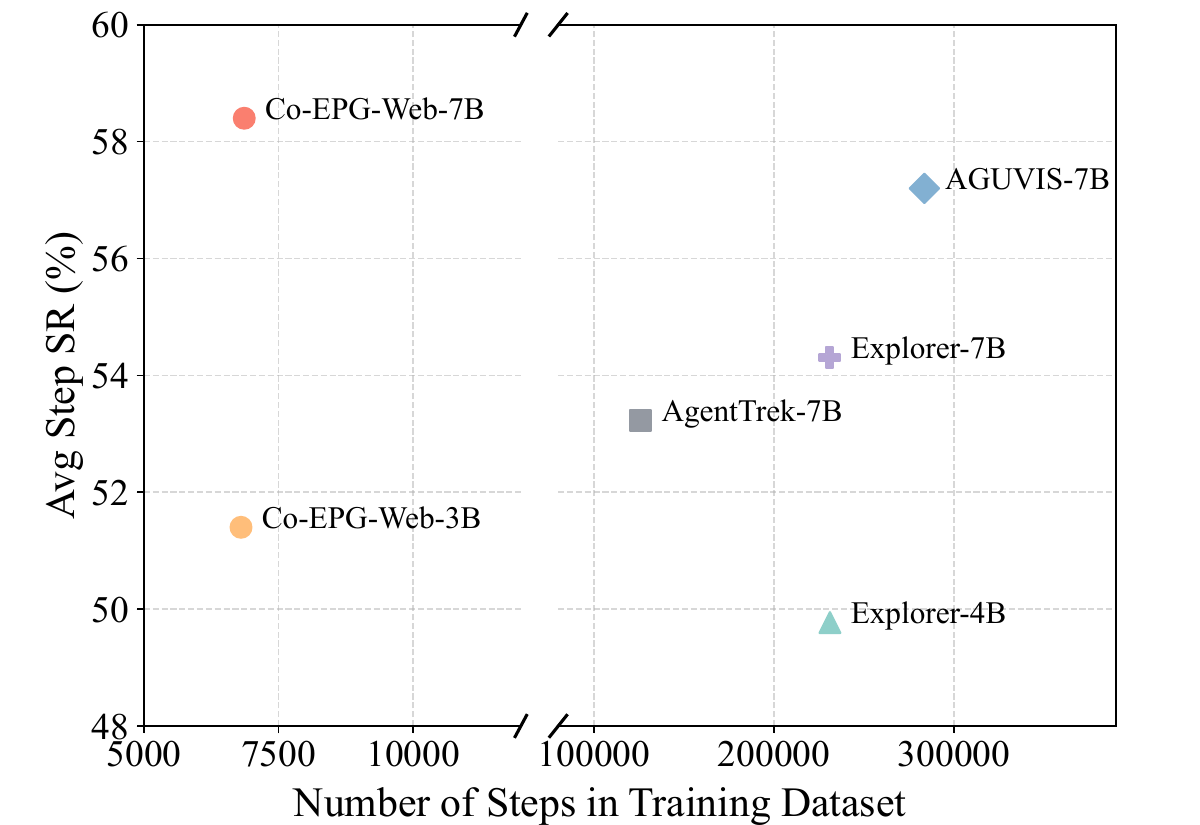}
    \caption{Performance and data efficiency on the Multimodal-Mind2Web benchmark. }
    \label{fig:experiment_1_1}
\end{figure}

\noindent\textbf{Data Efficiency.} Figure~\ref{fig:experiment_1_1} powerfully demonstrates the superior data efficiency of our Co-EPG framework. Remarkably, our Co-EPG-Web-7B surpasses the previous state-of-the-art model AGUVIS-7B on average Step SR while utilizing only 2.42\% of the labeled step data (6862 vs. 283500). This achievement validates the effectiveness of Co-EPG's innovative data value mining mechanism, particularly highlighting its advantages in low-resource scenarios.

\begin{figure}[ht]
    \centering
    \includegraphics[width=0.9\columnwidth]{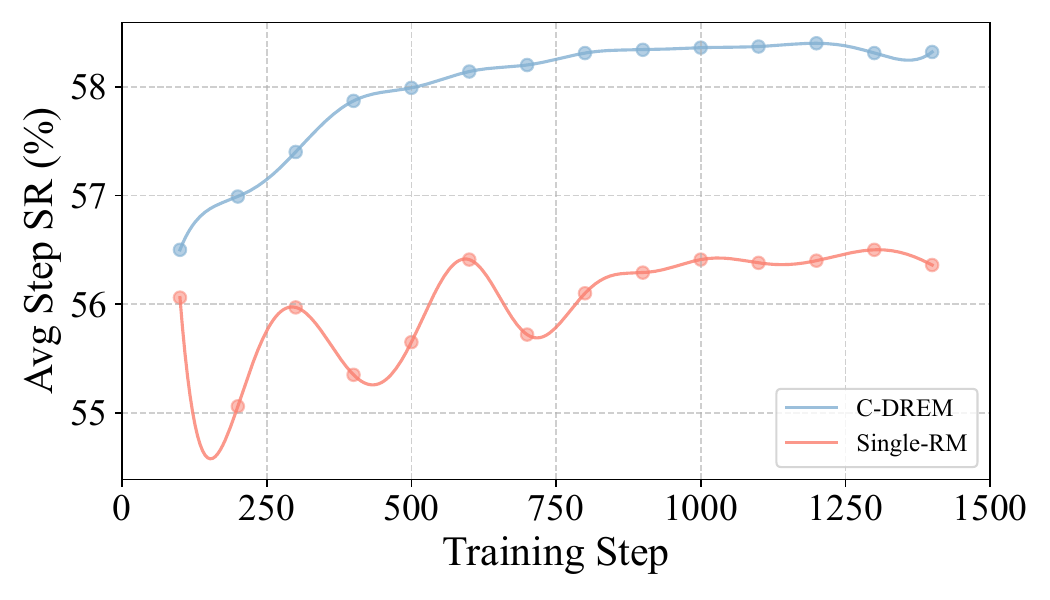}
    \caption{The performance curves of different reward mechanisms evaluated on Multimodal-Mind2Web.}
    \label{fig:analysis_3_1}
\end{figure}

\noindent\textbf{C-DREM Efficiency.}
On the Multimodal-Mind2Web dataset, we track Step SR to compare the efficiency of the C-DREM with a single grounding model. In Figure~\ref{fig:analysis_3_1}, our C-DREM plays a crucial role in enhancing learning stability and accelerating convergence efficiency. Relying on a single grounding model inevitably leads to cognitive blind spots, resulting in high-variance rewards during the agent's exploration process. This variability causes significant policy fluctuations and inefficient learning. In contrast, our proposed mechanism mitigates noise, enhances exploration efficiency, and speeds up the planning model's convergence by integrating multiple grounding models. 

\subsection{Analysis}
\begin{figure}[!ht]
    \centering
\includegraphics[width=0.9\columnwidth]{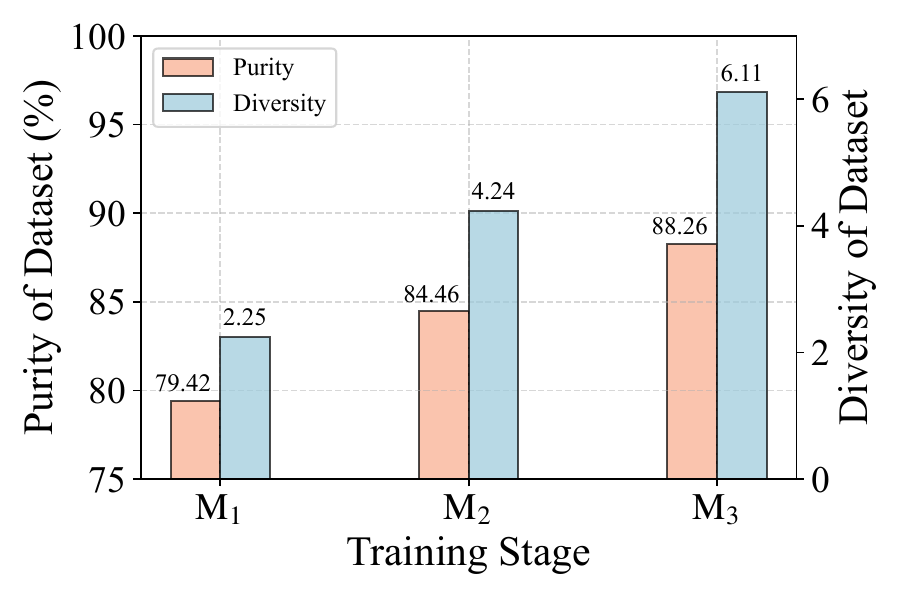}
    \caption{Iterative enhancement of data quality in Co-EPG.}
    \label{fig:analysis_1_1}
\end{figure}
\noindent\textbf{Evolution of Data Quality.} 
We evaluate the data quality evolution of the Co-EPG across two dimensions: purity and diversity. The purity metric is the proportion of plans successfully executed by grounding models, and the diversity metric is the average number of generated plans per task. As shown in Figure~\ref{fig:analysis_1_1}, both core metrics exhibit significant upward trends as the iteration progresses, with purity improving by 8.84\% and the diversity metric increasing by nearly 4. This result confirms the framework's data self-enhancement and highlights its significant potential for self-evolution.

\begin{figure}[!ht]
    \centering   \includegraphics[width=0.9\columnwidth]{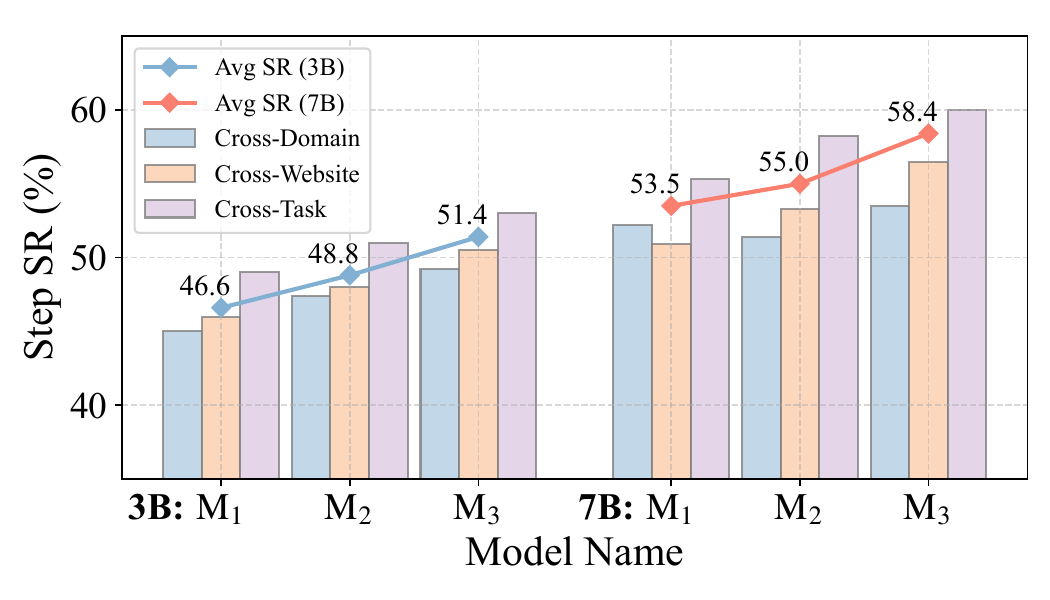}
    \caption{Iterative improvement of Co-EPG-Web models across all tasks on the Multimodal-Mind2Web.}
    \label{fig:experiment_1_2}
\end{figure}
\noindent\textbf{Improvement across Different Tasks.}
Figure~\ref{fig:experiment_1_2} demonstrates that on the Multimodal-Mind2web benchmark, both our 3B and 7B models exhibit significant and stable Step SR growth across all tasks. This steady improvement validates the robustness of our co-evolutionary learning process. Notably, this trend holds even for challenging cross-website and cross-domain tasks, demonstrating the framework's ability to foster strong generalization rather than mere memorization. Additionally, the consistent performance across various model scales demonstrates the effectiveness of Co-EPG.

\section{Conclusion}
In this paper, we propose Co-EPG, a self-iterative training framework for Co-Evolution of Planning and Grounding. Co-EPG utilizes planning instructions as interaction media, successfully achieving collaborative GRPO training between planning and grounding models. Extensive experiments demonstrate that Co-EPG exhibits state-of-the-art performance across both web and mobile tasks, showing stable performance improvement through iterative rounds. We believe that applying Co-EPG with other data synthesis techniques could unlock even greater data potential, and we hope our work can inspire future research in GUI agents.

\section*{Acknowledgments}
This work was supported by JST CREST Grant Number JPMJCR21M2, including the AIP Challenge Program.

\bibliography{aaai2026}

@article{OpenAI,
  title={Introducing deep research},
author={OpenAI},
  journal={https://openai.com/index/
introducing-deep-research/.},
year={2025}
}

@article{li2025websailor,
  title={WebSailor: Navigating Super-human Reasoning for Web Agent},
  author={Li, Kuan and Zhang, Zhongwang and Yin, Huifeng and Zhang, Liwen and Ou, Litu and Wu, Jialong and Yin, Wenbiao and Li, Baixuan and Tao, Zhengwei and Wang, Xinyu and others},
  journal={arXiv preprint arXiv:2507.02592},
  year={2025}
}

@article{zheng2024gpt,
  title={Gpt-4v (ision) is a generalist web agent, if grounded},
  author={Zheng, Boyuan and Gou, Boyu and Kil, Jihyung and Sun, Huan and Su, Yu},
  journal={arXiv preprint arXiv:2401.01614},
  year={2024}
}

@article{zhang2025litewebagent,
  title={Litewebagent: The open-source suite for vlm-based web-agent applications},
  author={Zhang, Danqing and Rama, Balaji and Ni, Jingyi and He, Shiying and Zhao, Fu and Chen, Kunyu and Chen, Arnold and Cao, Junyu},
  journal={arXiv preprint arXiv:2503.02950},
  year={2025}
}

@article{xu2024agenttrek,
  title={Agenttrek: Agent trajectory synthesis via guiding replay with web tutorials},
  author={Xu, Yiheng and Lu, Dunjie and Shen, Zhennan and Wang, Junli and Wang, Zekun and Mao, Yuchen and Xiong, Caiming and Yu, Tao},
  journal={arXiv preprint arXiv:2412.09605},
  year={2024}
}

@article{pahuja2025explorer,
  title={Explorer: Scaling exploration-driven web trajectory synthesis for multimodal web agents},
  author={Pahuja, Vardaan and Lu, Yadong and Rosset, Corby and Gou, Boyu and Mitra, Arindam and Whitehead, Spencer and Su, Yu and Awadallah, Ahmed},
  journal={arXiv preprint arXiv:2502.11357},
  year={2025}
}

@article{xu2024aguvis,
  title={Aguvis: Unified pure vision agents for autonomous gui interaction},
  author={Xu, Yiheng and Wang, Zekun and Wang, Junli and Lu, Dunjie and Xie, Tianbao and Saha, Amrita and Sahoo, Doyen and Yu, Tao and Xiong, Caiming},
  journal={arXiv preprint arXiv:2412.04454},
  year={2024}
}

@article{qi2024webrl,
  title={Webrl: Training llm web agents via self-evolving online curriculum reinforcement learning},
  author={Qi, Zehan and Liu, Xiao and Iong, Iat Long and Lai, Hanyu and Sun, Xueqiao and Zhao, Wenyi and Yang, Yu and Yang, Xinyue and Sun, Jiadai and Yao, Shuntian and others},
  journal={arXiv preprint arXiv:2411.02337},
  year={2024}
}

@article{luo2025gui,
  title={Gui-r1: A generalist r1-style vision-language action model for gui agents},
  author={Luo, Run and Wang, Lu and He, Wanwei and Xia, Xiaobo},
  journal={arXiv preprint arXiv:2504.10458},
  year={2025}
}

@article{wei2025webagent,
  title={Webagent-r1: Training web agents via end-to-end multi-turn reinforcement learning},
  author={Wei, Zhepei and Yao, Wenlin and Liu, Yao and Zhang, Weizhi and Lu, Qin and Qiu, Liang and Yu, Changlong and Xu, Puyang and Zhang, Chao and Yin, Bing and others},
  journal={arXiv preprint arXiv:2505.16421},
  year={2025}
}

@article{bai2023qwen,
  title={Qwen technical report},
  author={Bai, Jinze and Bai, Shuai and Chu, Yunfei and Cui, Zeyu and Dang, Kai and Deng, Xiaodong and Fan, Yang and Ge, Wenbin and Han, Yu and Huang, Fei and others},
  journal={arXiv preprint arXiv:2309.16609},
  year={2023}
}

@article{achiam2023gpt,
  title={Gpt-4 technical report},
  author={Achiam, Josh and Adler, Steven and Agarwal, Sandhini and Ahmad, Lama and Akkaya, Ilge and Aleman, Florencia Leoni and Almeida, Diogo and Altenschmidt, Janko and Altman, Sam and Anadkat, Shyamal and others},
  journal={arXiv preprint arXiv:2303.08774},
  year={2023}
}

@article{liu2024deepseek,
  title={Deepseek-v3 technical report},
  author={Liu, Aixin and Feng, Bei and Xue, Bing and Wang, Bingxuan and Wu, Bochao and Lu, Chengda and Zhao, Chenggang and Deng, Chengqi and Zhang, Chenyu and Ruan, Chong and others},
  journal={arXiv preprint arXiv:2412.19437},
  year={2024}
}

@article{wang2024qwen2,
  title={Qwen2-vl: Enhancing vision-language model's perception of the world at any resolution},
  author={Wang, Peng and Bai, Shuai and Tan, Sinan and Wang, Shijie and Fan, Zhihao and Bai, Jinze and Chen, Keqin and Liu, Xuejing and Wang, Jialin and Ge, Wenbin and others},
  journal={arXiv preprint arXiv:2409.12191},
  year={2024}
}

@article{yang2023set,
  title={Set-of-mark prompting unleashes extraordinary visual grounding in gpt-4v},
  author={Yang, Jianwei and Zhang, Hao and Li, Feng and Zou, Xueyan and Li, Chunyuan and Gao, Jianfeng},
  journal={arXiv preprint arXiv:2310.11441},
  year={2023}
}

@article{lu2024omniparser,
  title={Omniparser for pure vision based gui agent},
  author={Lu, Yadong and Yang, Jianwei and Shen, Yelong and Awadallah, Ahmed},
  journal={arXiv preprint arXiv:2408.00203},
  year={2024}
}

@article{deng2023mind2web,
  title={Mind2web: Towards a generalist agent for the web},
  author={Deng, Xiang and Gu, Yu and Zheng, Boyuan and Chen, Shijie and Stevens, Sam and Wang, Boshi and Sun, Huan and Su, Yu},
  journal={Advances in Neural Information Processing Systems},
  volume={36},
  pages={28091--28114},
  year={2023}
}

@article{gur2023real,
  title={A real-world webagent with planning, long context understanding, and program synthesis},
  author={Gur, Izzeddin and Furuta, Hiroki and Huang, Austin and Safdari, Mustafa and Matsuo, Yutaka and Eck, Douglas and Faust, Aleksandra},
  journal={arXiv preprint arXiv:2307.12856},
  year={2023}
}

@inproceedings{zhao2024expel,
  title={Expel: Llm agents are experiential learners},
  author={Zhao, Andrew and Huang, Daniel and Xu, Quentin and Lin, Matthieu and Liu, Yong-Jin and Huang, Gao},
  booktitle={Proceedings of the AAAI Conference on Artificial Intelligence},
  volume={38},
  number={17},
  pages={19632--19642},
  year={2024}
}

@article{fu2024autoguide,
  title={Autoguide: Automated generation and selection of context-aware guidelines for large language model agents},
  author={Fu, Yao and Kim, Dong-Ki and Kim, Jaekyeom and Sohn, Sungryull and Logeswaran, Lajanugen and Bae, Kyunghoon and Lee, Honglak},
  journal={Advances in Neural Information Processing Systems},
  volume={37},
  pages={119919--119948},
  year={2024}
}

@article{niu2024screenagent,
  title={Screenagent: A vision language model-driven computer control agent},
  author={Niu, Runliang and Li, Jindong and Wang, Shiqi and Fu, Yali and Hu, Xiyu and Leng, Xueyuan and Kong, He and Chang, Yi and Wang, Qi},
  journal={arXiv preprint arXiv:2402.07945},
  year={2024}
}

@article{lee2025learning,
  title={Learning to contextualize web pages for enhanced decision making by LLM agents},
  author={Lee, Dongjun and Lee, Juyong and Kim, Kyuyoung and Tack, Jihoon and Shin, Jinwoo and Teh, Yee Whye and Lee, Kimin},
  journal={arXiv preprint arXiv:2503.10689},
  year={2025}
}

@article{wu2024atlas,
  title={Os-atlas: A foundation action model for generalist gui agents},
  author={Wu, Zhiyong and Wu, Zhenyu and Xu, Fangzhi and Wang, Yian and Sun, Qiushi and Jia, Chengyou and Cheng, Kanzhi and Ding, Zichen and Chen, Liheng and Liang, Paul Pu and others},
  journal={arXiv preprint arXiv:2410.23218},
  year={2024}
}

@article{zhang2025symbiotic,
  title={Symbiotic cooperation for web agents: Harnessing complementary strengths of large and small llms},
  author={Zhang, Ruichen and Qiu, Mufan and Tan, Zhen and Zhang, Mohan and Lu, Vincent and Peng, Jie and Xu, Kaidi and Agudelo, Leandro Z and Qian, Peter and Chen, Tianlong},
  journal={arXiv preprint arXiv:2502.07942},
  year={2025}
}

@article{hui2025winclick,
  title={Winclick: Gui grounding with multimodal large language models},
  author={Hui, Zheng and Li, Yinheng and Chen, Tianyi and Banbury, Colby and Koishida, Kazuhito and others},
  journal={arXiv preprint arXiv:2503.04730},
  year={2025}
}

@article{gou2024navigating,
  title={Navigating the digital world as humans do: Universal visual grounding for gui agents},
  author={Gou, Boyu and Wang, Ruohan and Zheng, Boyuan and Xie, Yanan and Chang, Cheng and Shu, Yiheng and Sun, Huan and Su, Yu},
  journal={arXiv preprint arXiv:2410.05243},
  year={2024}
}

@article{fang2025webevolver,
  title={WebEvolver: Enhancing Web Agent Self-Improvement with Coevolving World Model},
  author={Fang, Tianqing and Zhang, Hongming and Zhang, Zhisong and Ma, Kaixin and Yu, Wenhao and Mi, Haitao and Yu, Dong},
  journal={arXiv preprint arXiv:2504.21024},
  year={2025}
}

@article{yuan2025enhancing,
  title={Enhancing Visual Grounding for GUI Agents via Self-Evolutionary Reinforcement Learning},
  author={Yuan, Xinbin and Zhang, Jian and Li, Kaixin and Cai, Zhuoxuan and Yao, Lujian and Chen, Jie and Wang, Enguang and Hou, Qibin and Chen, Jinwei and Jiang, Peng-Tao and others},
  journal={arXiv preprint arXiv:2505.12370},
  year={2025}
}

@article{zhao2025cola,
  title={COLA: A Scalable Multi-Agent Framework For Windows UI Task Automation},
  author={Zhao, Di and Ma, Longhui and Wang, Siwei and Wang, Miao and Lv, Zhao},
  journal={arXiv preprint arXiv:2503.09263},
  year={2025}
}

@inproceedings{chen2025guicourse,
  title={GUICourse: From General Vision Language Model to Versatile GUI Agent},
  author={Chen, Wentong and Cui, Junbo and Hu, Jinyi and Qin, Yujia and Fang, Junjie and Zhao, Yue and Wang, Chongyi and Liu, Jun and Chen, Guirong and Huo, Yupeng and others},
  booktitle={Proceedings of the 63rd Annual Meeting of the Association for Computational Linguistics (Volume 1: Long Papers)},
  pages={21936--21959},
  year={2025}
}

@article{zhang2024ufo,
  title={Ufo: A ui-focused agent for windows os interaction},
  author={Zhang, Chaoyun and Li, Liqun and He, Shilin and Zhang, Xu and Qiao, Bo and Qin, Si and Ma, Minghua and Kang, Yu and Lin, Qingwei and Rajmohan, Saravan and others},
  journal={arXiv preprint arXiv:2402.07939},
  year={2024}
}

@article{wang2024mobile,
  title={Mobile-agent-v2: Mobile device operation assistant with effective navigation via multi-agent collaboration},
  author={Wang, Junyang and Xu, Haiyang and Jia, Haitao and Zhang, Xi and Yan, Ming and Shen, Weizhou and Zhang, Ji and Huang, Fei and Sang, Jitao},
  journal={Advances in Neural Information Processing Systems},
  volume={37},
  pages={2686--2710},
  year={2024}
}

@article{liu2025infigui,
  title={Infigui-r1: Advancing multimodal gui agents from reactive actors to deliberative reasoners},
  author={Liu, Yuhang and Li, Pengxiang and Xie, Congkai and Hu, Xavier and Han, Xiaotian and Zhang, Shengyu and Yang, Hongxia and Wu, Fei},
  journal={arXiv preprint arXiv:2504.14239},
  year={2025}
}

@article{liu2025infiguiagent,
  title={Infiguiagent: A multimodal generalist gui agent with native reasoning and reflection},
  author={Liu, Yuhang and Li, Pengxiang and Wei, Zishu and Xie, Congkai and Hu, Xueyu and Xu, Xinchen and Zhang, Shengyu and Han, Xiaotian and Yang, Hongxia and Wu, Fei},
  journal={arXiv preprint arXiv:2501.04575},
  year={2025}
}

@article{qin2025ui,
  title={Ui-tars: Pioneering automated gui interaction with native agents},
  author={Qin, Yujia and Ye, Yining and Fang, Junjie and Wang, Haoming and Liang, Shihao and Tian, Shizuo and Zhang, Junda and Li, Jiahao and Li, Yunxin and Huang, Shijue and others},
  journal={arXiv preprint arXiv:2501.12326},
  year={2025}
}

@article{agashe2025agent,
  title={Agent s2: A compositional generalist-specialist framework for computer use agents},
  author={Agashe, Saaket and Wong, Kyle and Tu, Vincent and Yang, Jiachen and Li, Ang and Wang, Xin Eric},
  journal={arXiv preprint arXiv:2504.00906},
  year={2025}
}

@article{he2024webvoyager,
  title={Webvoyager: Building an end-to-end web agent with large multimodal models},
  author={He, Hongliang and Yao, Wenlin and Ma, Kaixin and Yu, Wenhao and Dai, Yong and Zhang, Hongming and Lan, Zhenzhong and Yu, Dong},
  journal={arXiv preprint arXiv:2401.13919},
  year={2024}
}

@article{cheng2024seeclick,
  title={Seeclick: Harnessing gui grounding for advanced visual gui agents},
  author={Cheng, Kanzhi and Sun, Qiushi and Chu, Yougang and Xu, Fangzhi and Li, Yantao and Zhang, Jianbing and Wu, Zhiyong},
  journal={arXiv preprint arXiv:2401.10935},
  year={2024}
}

@article{li2024effects,
  title={On the effects of data scale on ui control agents},
  author={Li, Wei and Bishop, William E and Li, Alice and Rawles, Christopher and Campbell-Ajala, Folawiyo and Tyamagundlu, Divya and Riva, Oriana},
  journal={Advances in Neural Information Processing Systems},
  volume={37},
  pages={92130--92154},
  year={2024}
}

@article{shao2024deepseekmath,
  title={Deepseekmath: Pushing the limits of mathematical reasoning in open language models},
  author={Shao, Zhihong and Wang, Peiyi and Zhu, Qihao and Xu, Runxin and Song, Junxiao and Bi, Xiao and Zhang, Haowei and Zhang, Mingchuan and Li, YK and Wu, Yang and others},
  journal={arXiv preprint arXiv:2402.03300},
  year={2024}
}

@article{qin2501ui,
  title={Ui-tars: Pioneering automated gui interaction with native agents, 2025},
  author={Qin, Yujia and Ye, Yining and Fang, Junjie and Wang, Haoming and Liang, Shihao and Tian, Shizuo and Zhang, Junda and Li, Jiahao and Li, Yunxin and Huang, Shijue and others},
  journal={URL https://arxiv. org/abs/2501.12326},
  year={2025}
}

@misc{zhao2024swiftascalablelightweightinfrastructure,
      title={SWIFT:A Scalable lightWeight Infrastructure for Fine-Tuning},
      author={Yuze Zhao and Jintao Huang and Jinghan Hu and Xingjun Wang and Yunlin Mao and Daoze Zhang and Zeyinzi Jiang and Zhikai Wu and Baole Ai and Ang Wang and Wenmeng Zhou and Yingda Chen},
      year={2024},
      eprint={2408.05517},
      archivePrefix={arXiv},
      primaryClass={cs.CL},
      url={https://arxiv.org/abs/2408.05517},
}

@inproceedings{rajbhandari2020zero,
  title={Zero: Memory optimizations toward training trillion parameter models},
  author={Rajbhandari, Samyam and Rasley, Jeff and Ruwase, Olatunji and He, Yuxiong},
  booktitle={SC20: International Conference for High Performance Computing, Networking, Storage and Analysis},
  pages={1--16},
  year={2020},
  organization={IEEE}
}

@article{dao2023flashattention,
  title={Flashattention-2: Faster attention with better parallelism and work partitioning, 2023},
  author={Dao, Tri},
  journal={URL https://arxiv. org/abs/2307.08691},
  year={2023}
}

@article{he2020deberta,
  title={Deberta: Decoding-enhanced bert with disentangled attention},
  author={He, Pengcheng and Liu, Xiaodong and Gao, Jianfeng and Chen, Weizhu},
  journal={arXiv preprint arXiv:2006.03654},
  year={2020}
}

@inproceedings{liu2025structural,
  title={Structural Reward Model: Enhancing Interpretability, Efficiency, and Scalability in Reward Modeling},
  author={Liu, Xiaoyu and Liang, Di and Shan, Hongyu and Liu, Peiyang and Liu, Yonghao and Wu, Muling and Li, Yuntao and Wu, Xianjie and Miao, Li and Shen, Jiangrong and others},
  booktitle={Proceedings of the 2025 Conference on Empirical Methods in Natural Language Processing: Industry Track},
  pages={672--685},
  year={2025}
}

@inproceedings{kapoor2024omniact,
  title={OmniACT: A Dataset and Benchmark for Enabling Multimodal Generalist Autonomous Agents for Desktop and Web},
  author={Kapoor, Raghav and Butala, Yash Parag and Russak, Melisa and Koh, Jing Yu and Kamble, Kiran and AlShikh, Waseem and Salakhutdinov, Ruslan},
  booktitle={European Conference on Computer Vision},
  pages={161--178},
  year={2024}
}

@misc{jiang2025importanceawaredataselectionefficient,
      title={Importance-Aware Data Selection for Efficient LLM Instruction Tuning}, 
      author={Tingyu Jiang and Shen Li and Yiyao Song and Lan Zhang and Hualei Zhu and Yuan Zhao and Xiaohang Xu and Kenjiro Taura and Hao Henry Wang},
      year={2025},
      eprint={2511.07074},
      archivePrefix={arXiv},
      primaryClass={cs.CL},
      url={https://arxiv.org/abs/2511.07074}, 
}

@article{wang2024large,
  title={Large language models as urban residents: An llm agent framework for personal mobility generation},
  author={Wang, Jiawei and Jiang, Renhe and Yang, Chuang and Wu, Zengqing and Onizuka, Makoto and Shibasaki, Ryosuke and Koshizuka, Noboru and Xiao, Chuan},
  journal={Advances in Neural Information Processing Systems},
  volume={37},
  pages={124547--124574},
  year={2024}
}

\clearpage

\appendix

\section{Datasets and Metrics} 
\label{sec:appendix Datasets and Metrics}
We mainly use two datasets:
\begin{itemize}
    \item \textbf{Multimodal-Mind2Web}~\cite{deng2023mind2web} is a multimodal extension of the web agent benchmark Mind2Web, offering a training set that includes 1,009 tasks and a test set comprising 1,013 tasks. The test set is meticulously divided into three categories: cross-task, cross-website, and cross-domain. These categories are ordered in increasing difficulty based on their differences from the training data distribution, aiming to rigorously evaluate the model's generalization performance at various levels. The action space of Multimodal-Mind2Web includes three primary operations: click, type, and select. For evaluation, we evaluate all test sets.
    \item \textbf{AndroidControl}~\cite{li2024effects} is a large-scale Android dataset whose training set encompasses 15,000 distinct tasks distributed across 833 applications, including both high-level and low-level tasks. Each instance of an AndroidControl task includes manually generated high-level and low-level instructions. We conduct the evaluation using a subset of 500 randomly sampled step-actions. We adhere to the standard data processing procedures outlined in~\cite{li2024effects}. During evaluation, the coordinates generated by the grounding models are mapped to the smallest visible element that encompasses them, similarly to ~\cite{gou2024navigating}.

\end{itemize}

\clearpage

\section{Implementation Details}
\label{sec:appendix Implementation Details}
We use MS-SWIFT~\cite{zhao2024swiftascalablelightweightinfrastructure} as the foundation for our training framework across all iterative stages, with extensive modifications to suit our specific tasks. For the planning model, we conduct SFT followed by GRPO training; for the grounding model, only SFT is performed.

We implement our method using Qwen2.5-VL-3B-Instruct and Qwen2.5-VL-7B-Instruct as the backbone models. For SFT on the planning model, we use a batch size of 96, a learning rate of $1 \times 10^{-6}$, a weight decay of 0.05, following a warmup plus cosine decay learning rate schedule. Training is performed for 3 epochs on 8 GPUs utilizing DeepSpeed~\cite{rajbhandari2020zero} ZeRO-3 for distributed training and FlashAttention-2~\cite{dao2023flashattention} to accelerate attention computation. For SFT on the grounding model, we adopt the same hyperparameters as the planning model. For GRPO training on the planning model, we utilize three distinct grounding models as reward sources (Qwen2.5-VL-32B-Instruct, Qwen2.5-VL-72B-Instruct, and our trained model $\phi_k$), with their static prior weights set to a ratio of 1:1:2 in C-DREM. We conduct the training for 3 epochs on 7 GPUs using DeepSpeed ZeRO-3 offload, which takes approximately 48 hours. The hyperparameters are configured as follows: a rollout number of 7 per group, a temperature of 0.9, a learning rate of $5 \times 10^{-7}$, and a batch size of 294. The KL divergence regularization coefficient ($\beta$) and the clip ratio ($\epsilon$) are configured to 0.01 and 0.2, respectively. To accelerate GRPO training, we optimize the framework to fetch rewards from all grounding models asynchronously, significantly reducing the time spent on this step. During both training and
inference, the height and width of the input images are resized to multiples of 28 before being fed into the model~\cite{bai2023qwen},  with a maximum pixel of $1024\times28\times28$. 

\clearpage

\section{Impact of Prior Weights Parameter on Model Performance}
\label{sec:appendix Impact of Prior Weights}
To further analyze the effect of static prior weights $\sigma$ in C-DREM on model performance, we conduct experiments by adjusting the prior weights assigned to the grounding models, which provide the rewards for plan generation. We designate our trained grounding model $\phi_k$ as a primary production model used for deployment inference, while Qwen2.5-VL-72B-Instruct and Qwen2.5-VL-32B-Instruct serve as additional VLMs. We experiment on the Qwen2.5-VL-7B-Instruct with several weight distributions that modulate the influence of the primary production model within the reward models. As shown in Table~\ref{tab:static_prior_table}, the results demonstrate that performance generally improves with an increased prior weight for the primary model. We observe that best performance is achieved with a prior weight ratio of 1:1:2 (Qwen2.5-VL-72B-Instruct: Qwen2.5-VL-32B-Instruct: Our Trained Model $\phi_k$), where the prior weight of the primary model is twice that of each auxiliary VLM. Increasing the weight futher to a 1:1:3 ratio lead to a slight decrease in performance, and overall improvements become marginal. Therefore, we choose the 1:1:2 configuration for the static prior weights in our main experiments, balancing reward stability with the importance of the primary model.

\begin{table*}[!ht]
    \centering
    \resizebox{\textwidth}{!}
    {
    \begin{tabular}{c cccccccccc}
    \toprule
    \multirow{2}{*}{\centering \textbf{Prior Weights}} & \multicolumn{3}{c}{\textbf{Cross-Task}} & \multicolumn{3}{c}{\textbf{Cross-Website}} & \multicolumn{3}{c}{\textbf{Cross-Domain}} & \multirow{2}{*}{\textbf{Avg SR}} \\
    \cmidrule(lr){2-4} \cmidrule(lr){5-7} \cmidrule(lr){8-10}
    & \textbf{Ele.Acc} & \textbf{Op.F1} & \textbf{Step SR} & \textbf{Ele.Acc} & \textbf{Op.F1} & \textbf{Step SR} & \textbf{Ele.Acc} & \textbf{Op.F1} & \textbf{Step SR} \\
    \midrule
    \multicolumn{1}{l}{\quad 1:1:1 } & \underline{66.0} & 90.9 & 60.7 & 61.2 & 90.3 & 56.3 & 59.0 & 91.1 & 55.0 & 57.3 \\
    \multicolumn{1}{l}{\quad 1:1:1.5} & 65.5 & \underline{91.1} & 60.7 & \textbf{62.9} & 91.0 & 57.4 & 59.2 & 91.1 & \textbf{55.5} & 57.9 \\
    \multicolumn{1}{l}{\quad 1:1:2} & \textbf{66.3} & \textbf{92.4} & \textbf{61.9} & \underline{62.3} & \textbf{91.7} & \textbf{58.1} & \underline{59.3} & \textbf{92.2} & 55.3 & \textbf{58.4} \\  
    \multicolumn{1}{l}{\quad 1:1:3} & 65.5 & 91.0 & \underline{60.9} & 61.8 & \underline{91.2} & \underline{57.6} & \textbf{59.4} & \underline{91.5} & \underline{55.4} & \underline{58.0} \\
    \bottomrule
    \end{tabular}
    }
    \caption{Impact of prior weights of (Qwen2.5-VL-72B-Instruct: Qwen2.5-VL-32B-Instruct: Our Trained
Model $\phi_k$) on model performance. We bold the best
results and underline the second-best performance.
    }
    \label{tab:static_prior_table}
\end{table*}

\clearpage

\section{Detailed Main Results}
\label{sec:appendix Detailed Main Results}
We conduct three iterative experiments on the Multimodal-Mind2Web and AndroidControl benchmarks, respectively. In each iteration, we report the performance of SFT alone (w/o GRPO) and the performance after applying GRPO. Table~\ref{tab:mm_mind2web_main_detail} presents the results for Multimodal-Mind2Web, and Table~\ref{tab:androidcontrol_main_detail} for AndroidControl. Our Co-EPG-Web-3B/7B and Co-EPG-Mob-3B/7B demonstrate significant and stable performance improvements across multiple iterations, confirming the performance enhancements brought by our self-evolution Co-EPG framework. Similarly, models trained solely with SFT demonstrate gradual performance improvements across training iterations. Therefore, this further validates the superiority of the self-evolution mechanism in our Co-EPG framework for enhancing data quality.

\begin{table*}[!ht]
    \centering
    \resizebox{\textwidth}{!}
    {
    \begin{tabular}{c cccccccccc}
    \toprule
    \multirow{2}{*}{\textbf{Training Stage}} & \multicolumn{3}{c}{\textbf{Cross-Task}} & \multicolumn{3}{c}{\textbf{Cross-Website}} & \multicolumn{3}{c}{\textbf{Cross-Domain}} & \multirow{2}{*}{\textbf{Avg SR}} \\
    \cmidrule(lr){2-4} \cmidrule(lr){5-7} \cmidrule(lr){8-10}
    & \textbf{Ele.Acc} & \textbf{Op.F1} & \textbf{Step SR} & \textbf{Ele.Acc} & \textbf{Op.F1} & \textbf{Step SR} & \textbf{Ele.Acc} & \textbf{Op.F1} & \textbf{Step SR} \\
    \midrule

    \multicolumn{1}{l}{Co-EPG-Web-3B} &  &  &  &  &  &  &  &  &  &  \\
    \multicolumn{1}{l}{\quad Iteration 1 w/o GRPO } & 52.4 & 87.1 & 47.5 & 48.7 & 85.8 & 43.1 & 47.7 & 87.2 & 44.0 & 44.9 \\
    \multicolumn{1}{l}{\quad Iteration 1 ($M_{1}$)} & 53.9 & 86.5 & 49.0 & 50.8 & 86.0 & 46.0 & 48.4 & \underline{88.1} & 45.0 & 46.6 \\

    \multicolumn{1}{l}{\quad Iteration 2 w/o GRPO } & 55.2 & 86.5 & 50.6 & 50.6 & \underline{87.0} & 46.1 & 48.3 & 87.8 & 45.0 & 47.2 \\
    \multicolumn{1}{l}{\quad Iteration 2 ($M_{2}$)} & 55.8 & 86.2 & 51.0 & 53.0 & 86.9 & \underline{48.0} & 51.5 & 87.1 & 47.4 & 48.8 \\

    \multicolumn{1}{l}{\quad Iteration 3 w/o GRPO } & \underline{56.3} & \underline{87.3} & \underline{51.3} & \underline{53.6} & 86.8 & 47.9 & \underline{51.9} & 87.8 & \underline{47.7} & \underline{49.0} \\
    \multicolumn{1}{l}{\quad Iteration 3 ($M_{3}$)} & \textbf{57.7} & \textbf{88.8} & \textbf{53.1} & \textbf{56.9} & \textbf{87.7} & \textbf{51.1} & \textbf{53.6} & \textbf{89.6} & \textbf{50.0} & \textbf{51.4} \\ \midrule

    \multicolumn{1}{l}{Co-EPG-Web-7B} &  &  &  &  &  &  &  &  &  &  \\
    
    \multicolumn{1}{l}{\quad Iteration 1 w/o GRPO } & 61.8 & 89.5 & 55.9 & 56.9 & 89.9 & 51.4 & 56.2 & 89.2 & 50.5 & 52.6 \\
    \multicolumn{1}{l}{\quad Iteration 1 ($M_{1}$)} & 61.6 & 89.0 & 56.0 & 56.4 & 90.2 & 51.5 & 57.2 & 90.7 & 53.1 & 53.5 \\
    
    \multicolumn{1}{l}{\quad Iteration 2 w/o GRPO } & 62.4 & 89.8 & 57.4 & 58.4 & 90.5 & 53.3 & 57.4 & 90.7 & 53.0 & 54.5 \\
    \multicolumn{1}{l}{\quad Iteration 2 ($M_{2}$)} & 64.5 & 90.1 & 58.9 & 58.8 & \underline{90.6} & 54.0 & 55.8 & 90.7 & 52.2 & 55.0 \\
    
    \multicolumn{1}{l}{\quad Iteration 3 w/o GRPO } & \underline{64.6} & \underline{90.1} & \underline{60.1} & \underline{60.5} & 90.0 & \underline{55.3} & \underline{58.2} & \underline{90.8} & \underline{54.1} & \underline{56.5} \\
    \multicolumn{1}{l}{\quad Iteration 3 ($M_{3}$)} & \textbf{66.3} & \textbf{92.4} & \textbf{61.9} & \textbf{62.3} & \textbf{91.7} & \textbf{58.1} & \textbf{59.3} & \textbf{92.2} & \textbf{55.3} & \textbf{58.4} \\    
    
    \bottomrule
    \end{tabular}
    }
    \caption{Detailed experimental results across three iterations on Multimodal-Mind2Web. We bold the best
results and underline the second-best performance.
    }
    \label{tab:mm_mind2web_main_detail}
\end{table*}

\clearpage

\begin{table}[!ht]
\centering
\small
    \begin{tabular}{c ccc c}
    \toprule
    \multirow{2}{*}{\textbf{Training Stage}} & \multicolumn{2}{c}{\textbf{Step Acc}} & \multirow{2}{*}{\textbf{Avg Acc}} \\
    \cmidrule(lr){2-3} 
    & \textbf{High} & \textbf{Low} & \\
    \midrule
    \multicolumn{1}{l}{Co-EPG-Mob-3B} & & &   \\   
    \multicolumn{1}{l}{\quad Iteration 1 w/o GRPO } & 61.2 & 87.6 & 74.4 \\   
    \multicolumn{1}{l}{\quad Iteration 1 ($M_{1}$)} & 63.8 & 87.8 & 75.8 \\   
    \multicolumn{1}{l}{\quad Iteration 2 w/o GRPO } & 64.6 & 89.4 & 77.0 \\   
    \multicolumn{1}{l}{\quad Iteration 2 ($M_{2}$)} & 68.4 & 89.6 & 79.0 \\   
    \multicolumn{1}{l}{\quad Iteration 3 w/o GRPO } & \underline{71.0} & \underline{90.0} & \underline{80.5} \\   
    \multicolumn{1}{l}{\quad Iteration 3 ($M_{3}$)} & \textbf{73.4} & \textbf{90.2} & \textbf{81.8} \\   \midrule
    
    \multicolumn{1}{l}{Co-EPG-Mob-7B} & & &   \\    
    \multicolumn{1}{l}{\quad Iteration 1 w/o GRPO } & 67.0 & 88.8 & 77.9 \\  
    \multicolumn{1}{l}{\quad Iteration 1 ($M_{1}$)} & 66.8 & 91.2 & 79.0 \\  
    \multicolumn{1}{l}{\quad Iteration 2 w/o GRPO } & 68.0 & 91.4 & 79.7 \\  
    \multicolumn{1}{l}{\quad Iteration 2 ($M_{2}$)} & 70.0 & 91.8 & 80.9 \\   
    \multicolumn{1}{l}{\quad Iteration 3 w/o GRPO } & \underline{72.0} & \underline{92.0} & \underline{82.0} \\   
    \multicolumn{1}{l}{\quad Iteration 3 ($M_{3}$)} & \textbf{74.2} & \textbf{92.0} & \textbf{83.1} \\   
   
    \bottomrule
    \end{tabular}
    \caption{
    Detailed experimental results across three iterations on AndroidControl. We bold the best
results and underline the second-best performance.
    }
    \label{tab:androidcontrol_main_detail}
\end{table}

\clearpage

\section{Results on More Environments}
\label{sec:appendix Results on More Environments}
To validate the generalization ability of Co-EPG, we conducted experiments in more diverse environments, such as on desktop applications. OmniACT ~\cite{kapoor2024omniact} is a cross-platform task dataset comprising 9,802 annotated instances, covering both desktop and web applications at a 3:1 ratio. The dataset spans native applications from MacOS (22), Linux (8), and Windows (8), along with 27 mainstream web applications, with 3-4 user interfaces annotated for each application. OmniACT is split into training, validation, and test sets in a 7:1:2 ratio. We train our models on the training set and evaluate on the test set. Consistent with DetACT ~\cite{kapoor2024omniact}, we adopt the Action Score as our evaluation metric.

The experimental results in Table ~\ref{tab:OmniACT_main} show that in the desktop and web environment, after two iterations,  Co-EPG-Des-7B-$\text{M}_2$ outperforms the previous state-of-the-art model, UGround-V1 \cite{gou2024navigating} by 19.2\%. This demonstrates generalization and effectiveness of the training framework across different platforms.

\begin{table}[!ht]
\centering
\small
    \begin{tabular}{ll ccc c}
    \toprule
    \textbf{Planner} & \textbf{Grounder} & \textbf{Avg Acc} \\
    \midrule
    GPT-4 & DetACT & 17.0 \\
    GPT-4 & SeeClick & 28.9 \\
    GPT-4 & UGround & 31.1 \\
    GPT-4o & SeeClick & 29.6 \\
    GPT-4o & UGround & 32.8 \\
    GPT-4o & UGround-V1-7B & \underline{34.0} \\
    \midrule
    \multicolumn{2}{c}{Co-EPG-Des-7B-$\text{M}_2$} & \textbf{53.2}  \\  
    \bottomrule
    \end{tabular}
    \caption{
    Performance on OmniACT. 
    We bold the best results and underline the second-best performance.
    }
    \label{tab:OmniACT_main}
\end{table}

\clearpage

\section{Prompt Templates}
\label{sec:appendix Prompt Templates}
Building upon the work presented in~\cite{gou2024navigating}, we implement subtle modifications to the prompt output format. Specifically, Table~\ref{tab:prompt_mm_planning} showcases the prompts designed for Multimodal-Mind2Web. Following the methodology outlined by ~\cite{deng2023mind2web}, we employ a pre-trained DeBERTa model ~\cite{he2020deberta} for candidate generation, from which we extract the top-50 highest-ranked elements to serve as referring elements in our framework. While for AndroidControl, the prompts in Table~\ref{tab:prompt_ac_planning_general} and Table~\ref{tab:prompt_ac_planning_guidelines} are referred to as the ``General Instruction`` prompt and ``Useful Guidelines`` prompt. Table~\ref{tab:prompt_ac_planning_task} provides the final, comprehensive prompt. Additionally, during the data enhancement process, to increase the likelihood of successful model verification of $\Phi$, we append an extra prompt section following the initial prompt, as illustrated in Table~\ref{tab:prompt_data_enhancement}.

\begin{table*}[!h]
\centering
\begin{tabular}{l}
\toprule
\textbf{Prompt Example} \\ \midrule
\textbf{[System Role]} \\ 
- You are imitating humans doing web navigation for a task step by step. \\
- At each stage, you can see the webpage like humans via a screenshot, and know the previous actions through recorded history. \\
- You need to decide on the next action to take. \\
- You can \textbf{CLICK} an element with the mouse, \textbf{SELECT} an option, or \textbf{TYPE} text with the keyboard. \\

\textbf{[Task Description]} \\
You are asked to complete the following task: \{task description\} \\

\textbf{[Previous Actions]} \\
\{previous actions\} \\
The screenshot below shows the current webpage. \\

\textbf{[Referring Elements]} \\
\{elements\} \\
From the screenshot, identify where and what each element is on the webpage. \\
Determine whether one matches your target element. Please examine the choices one by one. \\
If multiple options match, choose the most likely one by re-examining the screenshot, the choices, and your reasoning. \\

\textbf{[Useful Guidelines]} \\
First, observe the current webpage and think through your next step based on the task and previous actions. \\
To be successful, follow these rules: \\
1. Understand the task goal to avoid incorrect actions. \\
2. Carefully examine the current screenshot and issue a valid action based on observation. \\
3. You should only issue \textbf{one action at a time}. \\

Describe your thought process and reasoning in \textbf{three sentences}. \\

\textbf{[Output Format]} \\
Finally, conclude your answer using the format below. Ensure your output \textbf{strictly follows} this format: \\

\texttt{<plan>} \\
\hspace{1em} \textbf{ELEMENT}: Provide a description of the element you want to operate. \\
\texttt{</plan>} \\

\texttt{<action>} \\
\hspace{1em} \textbf{ACTION}: Choose one from \texttt{CLICK}, \texttt{TYPE}, or \texttt{SELECT}. Do not write ``None''. \\
\texttt{</action>} \\

\texttt{<value>} \\
\hspace{1em} \textbf{VALUE}: If \texttt{ACTION == TYPE}, specify the text to type. \\
\hspace{5em} If \texttt{ACTION == SELECT}, specify the option to be chosen. \\
\hspace{5em} Otherwise, write ``None''. \\
\texttt{</value>} \\

\bottomrule
\end{tabular}
\caption{The Prompt for Multimodal-Mind2Web.}
\label{tab:prompt_mm_planning}
\end{table*}

\clearpage

\begin{table*}[!h]
\centering
\begin{tabular}{l}
\toprule

\textbf{Prompt Example} \\
\midrule

\textbf{[General Instruction]} \\
- You are an agent who can operate an Android phone on behalf of a user. \\
- Based on the user's goal or request, you may complete tasks by performing actions step by step on the phone. \\
- For each step, you are given the current screenshot and a history of past actions. \\
- You must use this information and the user's goal to decide the next action. \\

\textbf{[Output Format]} \\
\texttt{<plan>} \\
\hspace{1em} \textbf{ELEMENT}: If you want to click or long press, describe the target element you want to operate. \\
\hspace{1em} Otherwise, write ``None''. \\
\texttt{</plan>} \\

\texttt{<action>} \\
\hspace{1em} \textbf{ACTION}: Choose an action from the following options: \\
\hspace{2em} \texttt{terminate}, \texttt{click}, \texttt{long\_press}, \texttt{input\_text}, \texttt{scroll}, \\
\hspace{2em} \texttt{navigate\_home}, \texttt{navigate\_back}, \texttt{open\_app}, \texttt{wait} \\
\hspace{2em} You must choose one of these, instead of choosing 
``None''. \\
\texttt{</action>} \\

\texttt{<value>} \\
\hspace{1em} \textbf{VALUE}: Provide additional input based on \textbf{ACTION}: \\
\hspace{2em} If \texttt{ACTION == input\_text}, specify the text. \\
\hspace{2em} If \texttt{ACTION == scroll}, specify direction from \texttt{up/down/left/right}. \\
\hspace{2em} If \texttt{ACTION == open\_app}, specify the app name. \\
\hspace{2em} Otherwise, write ``None''. \\
\texttt{</value>} \\

\textbf{[Action Descriptions]} \\
- \textbf{terminate}: Use if the task is complete or cannot be completed. \\
- \textbf{click}: Tap on an element. Describe it in ELEMENT. \\
- \textbf{long\_press}: Same as click, but long press. \\
- \textbf{input\_text}: Type into a field. Provide the input in VALUE. \\
- \textbf{scroll}: Scroll the screen. VALUE should be one of \texttt{up/down/left/right}. \\
- \textbf{navigate\_home}: Return to the home screen. \\
- \textbf{navigate\_back}: Go back one screen. \\
- \textbf{open\_app}: Launch an app. Specify app name in VALUE. \\
- \textbf{wait}: Wait for the screen to update. \\

\bottomrule
\end{tabular}
\caption{The General Instruction Prompt for AndroidControl.}
\label{tab:prompt_ac_planning_general}
\end{table*}

\clearpage

\begin{table*}[!h]

\centering
\begin{tabular}{p{0.96\textwidth}}
\toprule
\textbf{Useful Guidelines} \\

\midrule
Here are some useful guidelines you need to follow: \\
\\

\textbf{General:} \\ 
- Usually, there will be multiple ways to complete a task, pick the easiest one. Also, when something does not work as 
expected (due to various reasons), sometimes a simple retry can solve the problem, but if it doesn't (you can see that 
from the history), \textbf{SWITCH} to other solutions. \\
- If the desired state is already achieved (e.g., enabling Wi-Fi when it's already on), you can just complete the task. \\
\\

\textbf{Action Related:} \\
- Use the \texttt{open\_app} action whenever you want to open an app (nothing will happen if the app is not installed), do not use the app drawer to open an app unless all other ways have failed. \\
- Use the \texttt{input\_text} action whenever you want to type something (including password) instead of clicking characters on the keyboard one by one. Sometimes there is some default text in the text field you want to type in, remember to delete them before typing. \\
- For \texttt{click} and \texttt{long\_press}, the element you pick must be \textbf{VISIBLE} in the screenshot to interact with it.\\
- The \texttt{element} field requires a concise yet comprehensive description of the target element in a single sentence, not exceeding 30 words. Include all essential information to uniquely identify the element. If you find identical elements, specify their location and details to differentiate them from others. \\
- Consider exploring the screen by using the \texttt{scroll} action with different directions to reveal additional content. \\

- The direction parameter for the \texttt{scroll} action specifies the direction in which the content moves and opposites to swipe; for example, to view content at the bottom, the \texttt{scroll} direction should be set to \texttt{down}. \\
\\

\textbf{Text Related Operations:} \\
- Normally, to select certain text on the screen: \\ 
\textit{(i)} Enter text selection mode by long pressing the area where the text is, then some of the words near the long press point will be selected (highlighted with two pointers indicating the range), and usually a text selection bar  will also appear with options like \texttt{copy}, \texttt{paste}, \texttt{select all}, etc. \\
\textit{(ii)} Select the exact text you need. Usually, the text selected from the previous step is \textbf{NOT} the one you want, you need to adjust the range by dragging the two pointers. If you want to select all text in the text field, simply click the \texttt{select all} button in the bar. \\

\\

\textbf{Note:} \\
- You don't have the ability to drag something around the screen, so in general, you can not select arbitrary text. \\
- To delete some text: the most traditional way is to place the cursor at the right place and use the backspace button in the keyboard to delete the characters one by one (can long press the backspace to accelerate if there are many to delete). Another approach is to first select the text you want to delete, then click the backspace button on the keyboard. \\
- To copy some text: first select the exact text you want to copy, which usually also brings up the text selection bar, then click the \texttt{copy} button in the bar. \\
- To paste text into a text box, first long press the text box, then usually the text selection bar will appear with a \texttt{paste} button in it. \\
- When typing into a text field, sometimes an auto-complete dropdown list will appear. This usually indicates this is an enum field, and you should try to select the best match by clicking the corresponding one in the list. \\

\bottomrule
\end{tabular}
\caption{The Useful Guidelines Prompt for AndroidControl}
\label{tab:prompt_ac_planning_guidelines}
\end{table*}

\clearpage

\begin{table*}[!h]
\centering
\begin{tabular}{l}
\toprule
\textbf{Prompt Example} \\ \midrule

\textbf{[System Role]} \\
You are a helpful assistant. \\

\midrule
\textbf{[High-Level Prompt]} \\

\{General Instruction\} \\

The current user goal/request is: high-level task \\

Here is a history of what you have done so far: \\
\{history\} \\

The current raw screenshot is given to you. \\

\{Useful Guidelines\} \\
\texttt{additional\_guidelines:} \\

Now output an action from the above list in the correct format. Your answer should look like: \\
\texttt{<plan>} ELEMENT: ... \texttt{</plan>} \\
\texttt{<action>} ACTION: ... \texttt{</action>} \\
\texttt{<value>} VALUE: ... \texttt{</value>} \\

Your Answer: \\

\midrule
\textbf{[Low-Level Prompt]} \\

\{General Instruction\} \\

The current user goal/request is: high-level task \\

The current next step's low-level goal is: low-level task \\

Here is a history of what you have done so far: \\
\{history\} \\

The current raw screenshot is given to you. \\

\{Useful Guidelines\} \\
\texttt{additional\_guidelines:} \\

Now output an action from the above list in the correct format. Your answer should look like: \\
\texttt{<plan>} ELEMENT: ... \texttt{</plan>} \\
\texttt{<action>} ACTION: ... \texttt{</action>} \\
\texttt{<value>} VALUE: ... \texttt{</value>} \\

Your Answer: \\

\bottomrule
\end{tabular}
\caption{The High-Level Prompt and Low-Level Prompt for AndroidControl.}
\label{tab:prompt_ac_planning_task}
\end{table*}

\clearpage

\begin{table*}[!h]
\centering

\resizebox{\textwidth}{!}
{
\begin{tabular}{l}
\toprule
\textbf{Additional Prompt}   \\  
\midrule
Now, I will provide you with the correct ACTION and VALUE content, and correct the element bbox in the picture. \\
Please use this information to supplement the ELEMENT value and output in the specified format. \\
ACTION: \{action\} \\
VALUE: \{value\} \\
ELEMENT\_BBOX: \{target bbox\} \\

\bottomrule
\end{tabular}
}
\caption{The Additional Prompt in Data Enhancement for Multimodal-Mind2Web and AndroidControl.}
\label{tab:prompt_data_enhancement}
\end{table*}

\clearpage

\section{Evaluation Examples}
\label{sec:appendix G}
\begin{figure*}[!ht]
    \centering
    \includegraphics[width=\textwidth]{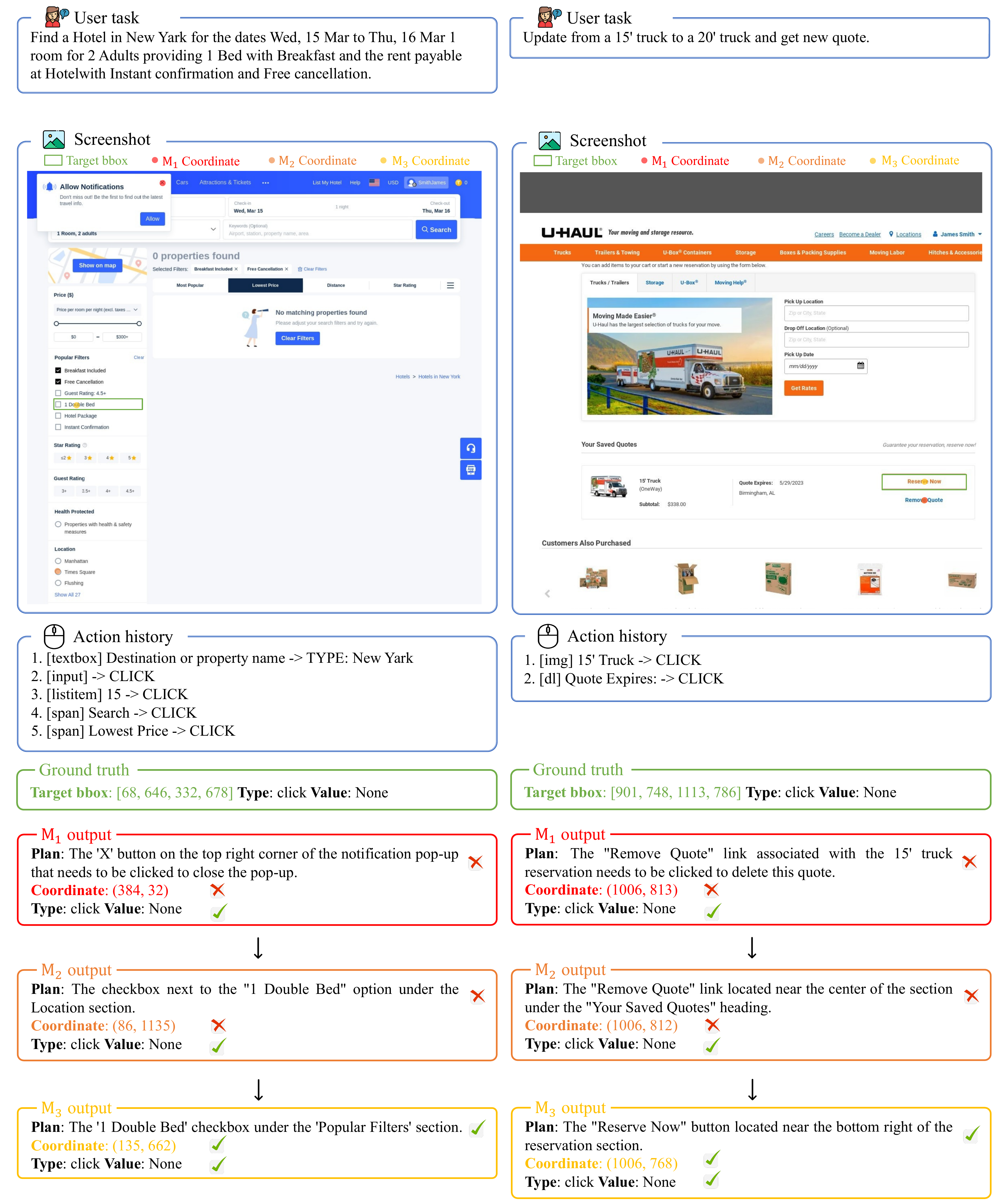}
    \caption{Evolution Examples of Co-EPG on Multimodal-Mind2Web.
    }
    \label{fig:app_case_mm_3}
\end{figure*}

\begin{figure*}[!ht]
    \centering
    \includegraphics[width=\textwidth]{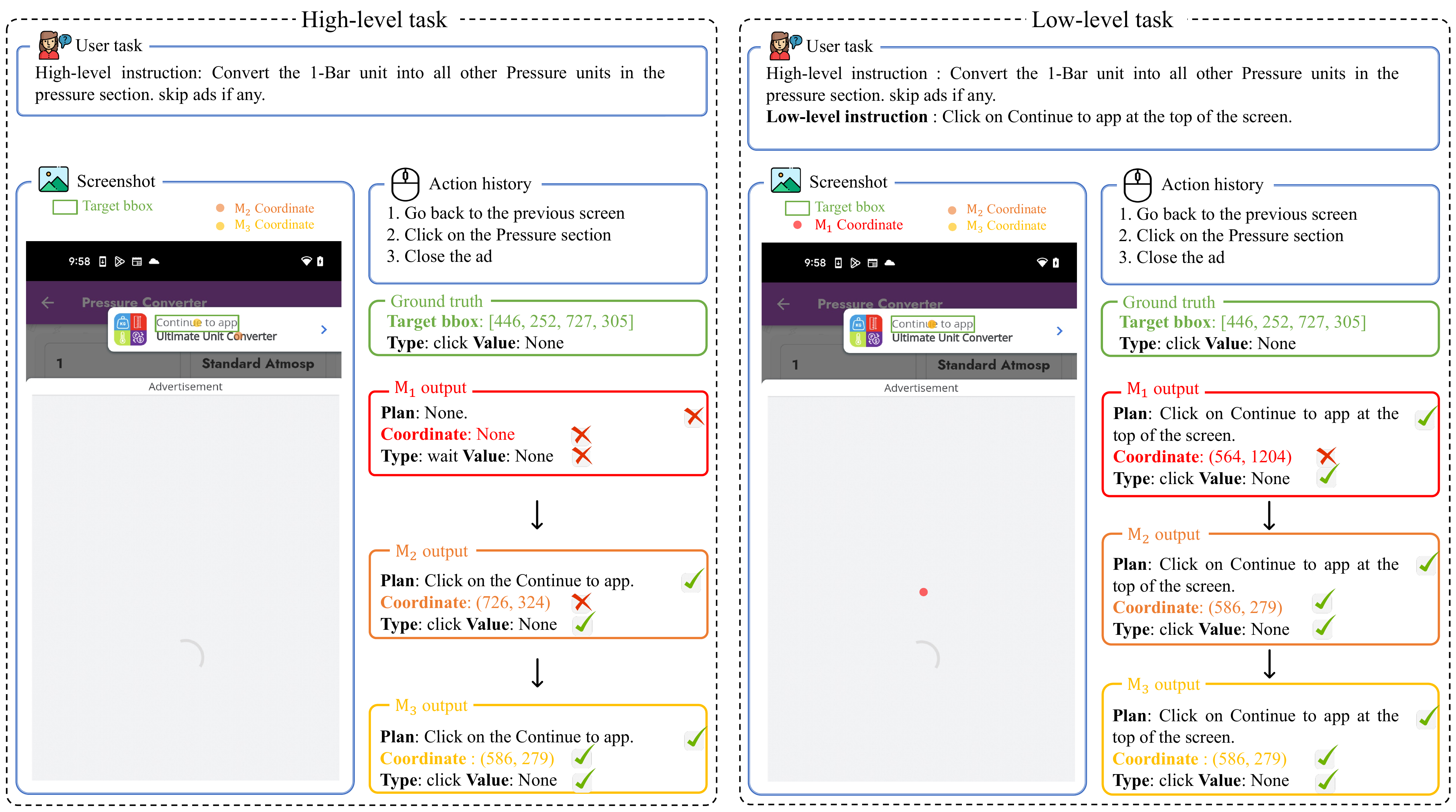}
    \caption{Evolution Examples of Co-EPG on AndroidControl.
    }
    \label{fig:app_case_ac}
\end{figure*}

Figure~\ref{fig:app_case_mm_3} presents the inference results of the Multimodal-Mind2Web test set samples under $M_1$, $M_2$, and $M_3$ in three iterative steps. Figure~\ref{fig:app_case_ac} shows the inference results of the AndroidControl test set samples under $M_1$, $M_2$, and $M_3$ in three iterative steps. It can be seen that as the number of iterations increases, the inference performance of both test sets gradually improves.

\clearpage

\end{document}